\documentclass[10pt,twocolumn,letterpaper]{article}

\usepackage[pagenumbers]{iccv} % To force page numbers, e.g. for an arXiv version
\usepackage{graphicx}
\usepackage{caption}
\usepackage{tabularx}
\usepackage{multirow}
\usepackage{algorithm}
\usepackage{algpseudocode}
\usepackage{amsmath}
\usepackage{amssymb}
\definecolor{iccvblue}{rgb}{0.21,0.49,0.74}
\usepackage[pagebackref,breaklinks,colorlinks,allcolors=iccvblue]{hyperref}

\def\paperID{1723} % *** Enter the Paper ID here
\def\confName{ICCV}
\def\confYear{2025}

\title{DreamFuse: Adaptive Image Fusion with Diffusion Transformer}

\author{
    \textbf{Junjia Huang}$^{1,2}$\thanks{Equal Contribution.}\quad \textbf{Pengxiang Yan}$^{2*}$\quad \textbf{Jiyang Liu}$^{2*}$\thanks{Project Lead.}\quad \textbf{Jie Wu}$^{2}$\\
    \textbf{Zhao Wang}$^2$\quad  \textbf{Yitong Wang}$^{2}$\quad  \textbf{Liang Lin}$^{1,3}$\quad  \textbf{Guanbin Li}$^{1,3,4}$\thanks{Corresponding Author.}\\
    {$^1$Sun Yat-sen University, $^2$ByteDance Intelligent Creation}, $^3$Peng Cheng Laboratory  \\
    $^4$Guangdong Key Laboratory of Big Data Analysis and Processing \\
    {\href{https://ll3rd.github.io/DreamFuse/}{https://ll3rd.github.io/DreamFuse/}}
}

\begin{document}

\twocolumn[{
  \maketitle
  \begin{center}
    \includegraphics[width=0.97\linewidth]{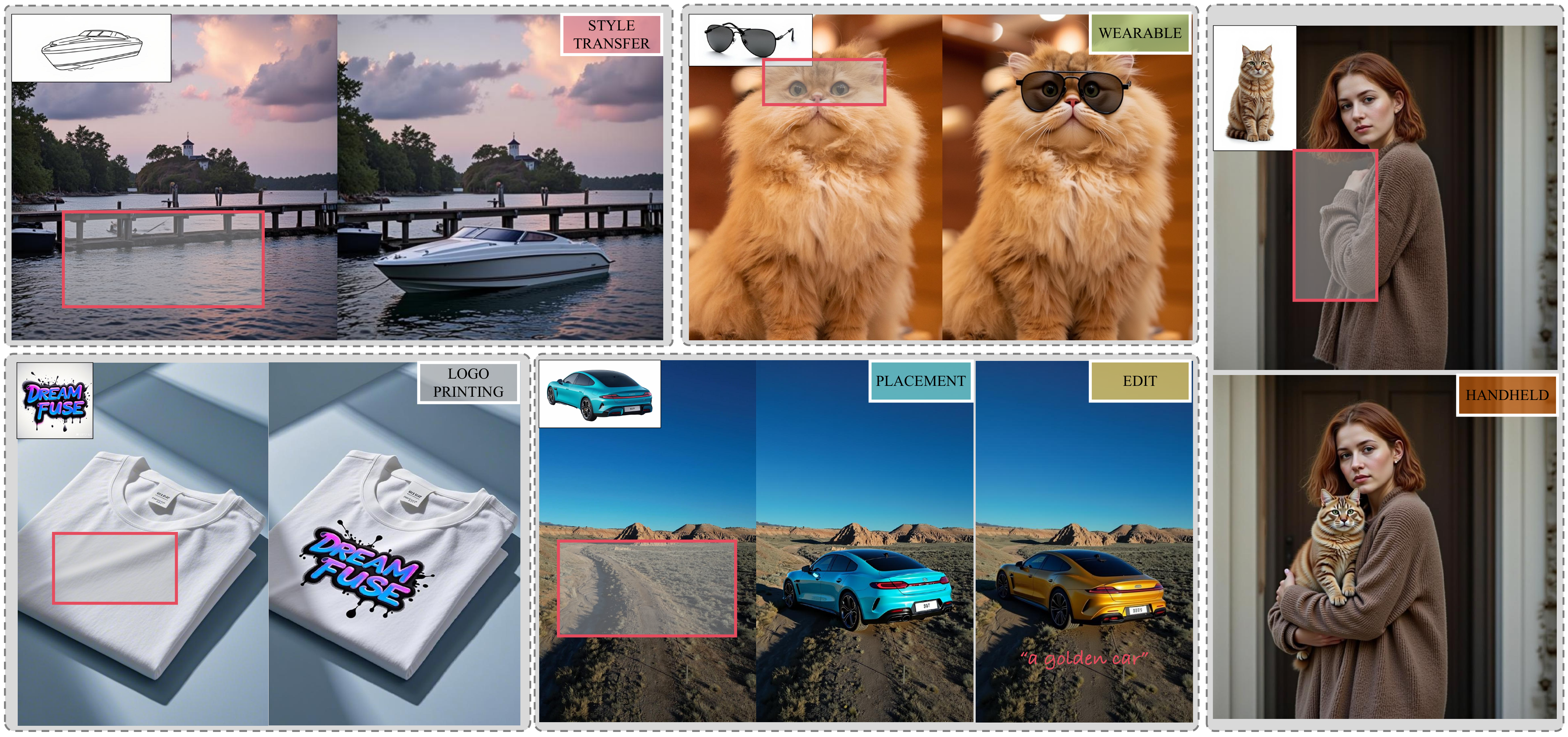}
    \captionof{figure}{DreamFuse demonstrates adaptive performance across diverse scenarios, including style transfer, wearable items, logo printing, placement and handheld. Notably, when given a text prompt, our method effectively responds by further editing the attributes of the foreground object (e.g., a golden car).}
    \label{fig: intro}
  \end{center}
}]

  \renewcommand{\thefootnote}{\fnsymbol{footnote}}  % 符号风格匹配 \thanks 的 *
  \footnotetext[1]{Equal Contribution.}
  \footnotetext[2]{Project Lead.}
  \footnotetext[3]{Corresponding Author.}
  
\begin{abstract}
Image fusion seeks to seamlessly integrate foreground objects with background scenes, producing realistic and harmonious fused images. Unlike existing methods that directly insert objects into the background, adaptive and interactive fusion remains a challenging yet appealing task. It requires the foreground to adjust or interact with the background context, enabling more coherent integration. To address this, we propose an iterative human-in-the-loop data generation pipeline, which leverages limited initial data with diverse textual prompts to generate fusion datasets across various scenarios and interactions, including placement, holding, wearing, and style transfer. Building on this, we introduce DreamFuse, a novel approach based on the Diffusion Transformer (DiT) model, to generate consistent and harmonious fused images with both foreground and background information. DreamFuse employs a Positional Affine mechanism to inject the size and position of the foreground into the background, enabling effective foreground-background interaction through shared attention. Furthermore, we apply Localized Direct Preference Optimization guided by human feedback to refine DreamFuse, enhancing background consistency and foreground harmony. DreamFuse achieves harmonious fusion while generalizing to text-driven attribute editing of the fused results.
Experimental results demonstrate that our method outperforms state-of-the-art approaches across multiple metrics.
\end{abstract}
\section{Introduction}
Foreground-background image fusion is a fundamental and practical task in image editing. Recently, with the rapid development of generation methods based on diffusion models, numerous innovative and imaginative approaches have emerged in this field. Beyond generating images purely driven by text prompts, increasing attention has been directed toward generating customized images guided by specific foreground  objects~\cite{ruiz2023dreambooth, chen2023subject, tan2024ominicontrol} or performing inpainting on designated background regions~\cite{yang2023paint} for image editing and fusion.
Going further, several methods aim to achieve harmonious fusion of foreground and background by adjusting details such as lighting~\cite{zhang2025scaling}, shadows~\cite{tarres2024thinking}, and brightness-contrast consistency~\cite{wang2023semi}, making the fused images appear more natural. Other approaches~\cite{chen2024anydoor, winter2024objectmate, he2024affordance} strive to enhance fusion by modifying the orientation, pose, or style of the foreground object while preserving its identity attributes, enabling better adaptation to the background. However, most of these methods typically focus on directly placing the foreground object into the background scene. In contrast, practical scenarios often involve more diverse and interactive cases, such as partial occlusions, alternating visibility, or interactions where the object is held, worn, or integrated into the scene.

A major challenge in handling such complex fusion scenarios is the lack of suitable datasets. Existing methods typically rely on object segmentation from images or videos~\cite{kirillov2023segment, miao2022large} followed by inpainting~\cite{rombach2022high, suvorov2022resolution} to reconstruct backgrounds. However, this multi-step process frequently suffers from quality degradation due to imprecise segmentation or suboptimal inpainting, which can introduce artifacts like shadows or residual elements in the background. Moreover, handling partially occluded foregrounds is particularly challenging~\cite{tudosiu2024mulan}, and segmentation-based data often fails to adjust object pose or perspective to align with the background during fusion.
Based on these observations, we propose an Iterative Human-in-the-Loop Data Generation Pipeline to directly generate fused data, including foregrounds, backgrounds, and the fused images, avoiding issues such as incomplete foregrounds and background artifacts. We train a DiT model on curated fused data with text prompts, modifying its attention mechanism to shared attention to ensure identity consistency across fused data. Using this model, we generate multi-scale fused data for various scenarios with diverse prompts, such as placement, handheld interactions, wearable items, logo printing, and style transfer. Throughout the process, we enhance content diversity by incorporating existing LoRA~\cite{hu2021lora} and iteratively optimize the model through manual selected data. We utilize GPT-4o to filter out low-quality fused data, such as mismatched foregrounds or degraded images, ultimately constructing a dataset of 80,000 high-quality, multi-scene, multi-scale fused image samples.

\begin{figure*}[t]
  \centering
   \includegraphics[width=0.95\linewidth]{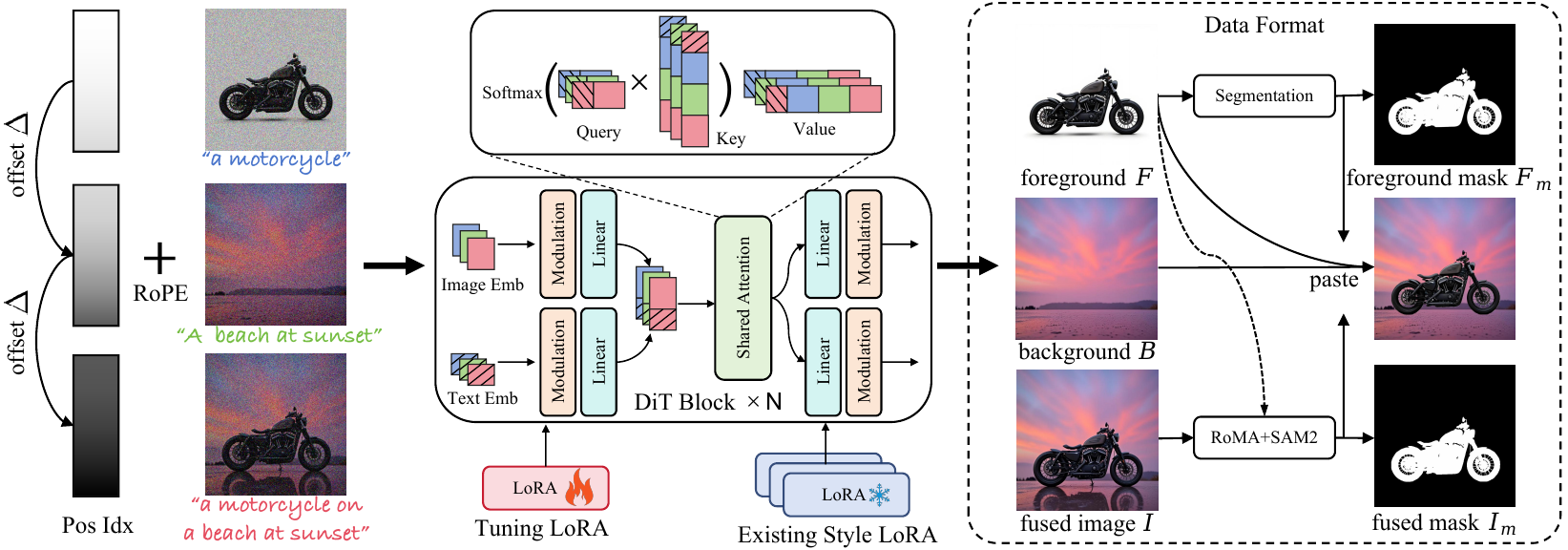}
   \caption{The framework of the data generation model and position matching process. The left side of the image illustrates the design structure of our data generation model, while the right side shows the position matching process and data format. We enhance the diversity of fused data generation through flexible and rich prompts combined with various style LoRAs.}
   \label{fig:data gen}
\end{figure*}

Another critical challenge in image fusion is ensuring background consistency and foreground harmony. Some approaches~\cite{chen2024anydoor, he2024affordance} rely on masks or bounding boxes for foreground placement, blending the background outside the mask to maintain consistency. However, these approaches often fail to realistically render effects like shadows or reflections beyond the mask, limiting the realism of the fused results. Other methods~\cite{lu2023tf, wang2024primecomposer} reconstruct fused images through inversion, improving foreground harmony but often compromising background consistency.
To address this trade-off, we propose DreamFuse, an adaptive image fusion framework based on DiT. By incorporating shared attention, we condition the the fused image generation on both the foreground and background while employing positional affine to introduce the foreground's position and scale without restricting its editable regions. Additionally, we employ Localized Direct Preference Optimization (LDPO) to further optimize the foreground and background regions of the fused image, ensuring better alignment with human preferences. Experimental results demonstrate that DreamFuse performs exceptionally well across various scenarios. As shown in \cref{fig: intro}, DreamFuse produces highly realistic fusion effects for real-world images. Furthermore, during training, a certain proportion of fused image descriptions is incorporated as prompts. When given a text prompt, DreamFuse effectively responds to the input and enables attribute modifications in the fused scenes, such as turning a car into gold.

In summary, our key contributions are threefold:
\begin{itemize}
    \item We propose an iterative Human-in-the-Loop data generation pipeline and construct a comprehensive fusion dataset containing 80k diverse fusion scenarios.
    \item We propose DreamFuse, a fusion framework based on DiT, which leverages positional affine and LDPO strategies to integrate the foreground into the background more naturally and adaptively.
    \item Our method outperforms the state-of-the-art methods on various benchmarks and remains effective in real-world and out-of-distribution scenarios.
\end{itemize}

\section{Related Work}
\noindent\textbf{\textit{Customized Image Generation.}} Customized image generation aims to create user-specific images based on text prompts or reference images. Some methods~\cite{han2023svdiff, ruiz2023dreambooth, gal2022image, kumari2023multi} incorporate reference concepts into specific text prompts, while approaches~\cite{ye2023ip, zhang2024ssr, wang2024instantid} utilize additional encoders to encode reference images as visual prompts, introducing customized representations for generation. Other methods~\cite{tan2024ominicontrol, huang2024context} achieve subject-driven generation by directly concatenating reference and target images during generation. In this paper, we further extend generative diffusion models by fine-tuning on small-scale data to directly generate customized foregrounds, backgrounds, and fused images based on corresponding text prompts.

\noindent\textbf{\textit{Image Fusion.}} The goal of image fusion is to seamlessly integrate an object from a foreground image into a background image. Compared to directly cutting and pasting, some approaches~\cite{peng2024frih, zhou2024foreground, menghigh, zhang2025scaling} adjust the lighting, shadows, and colors of the pasted foreground region to achieve a more harmonious fusion. Other methods~\cite{lu2023tf, wang2024primecomposer, tao2024motioncom, winter2024objectdrop, chen2024anydoor, he2024affordance, winter2024objectmate} focus on altering the perspective, pose, or style of the foreground object to make it fit more naturally into the background image. However, these methods are often restricted to object placement. In this paper, we propose a more versatile fusion approach that supports a variety of scenarios, including hand-held objects, wearable items, and style transformations, enabling more diverse integrations.

\noindent\textbf{\textit{Human Feedback Learning.}} Many methods now leverage human feedback learning to make generated images more aligned with users’ preferences. Some approaches~\cite{xu2023imagereward, xu2024visionreward, liu2025improving} train a reward model to understand human preferences and improve the generation quality through reward feedback learning. Others~\cite{wallace2024diffusion, li2024aligning, zhang2025diffusion} utilize human comparison data to directly optimize a policy that best satisfies human preferences. For the image fusion task, we propose localized direct preference optimization, which focuses on region-specific optimization to enhance both the background consistency and the harmony in fused images.

\section{Methodology}
As shown in the data format in \cref{fig:data gen}, the image fusion task typically involves the following types of images: a foreground image $F \in \mathbb{R}^{H\times W\times 3}$ with mask $F_m \in \mathbb{R}^{H\times W}$, a background image $B \in \mathbb{R}^{H\times W\times 3}$, a fused image $I \in \mathbb{R}^{H\times W\times 3}$, a fused mask $I_m \in \mathbb{R}^{H\times W}$ associated with the fused object. The masks $F_m$ and $I_m$ are primarily used to indicate the position and size of the foreground object. In practical applications, only the centroid and bounding box of the mask are required.

\begin{figure*}[t]
  \centering
   \includegraphics[width=1\linewidth]{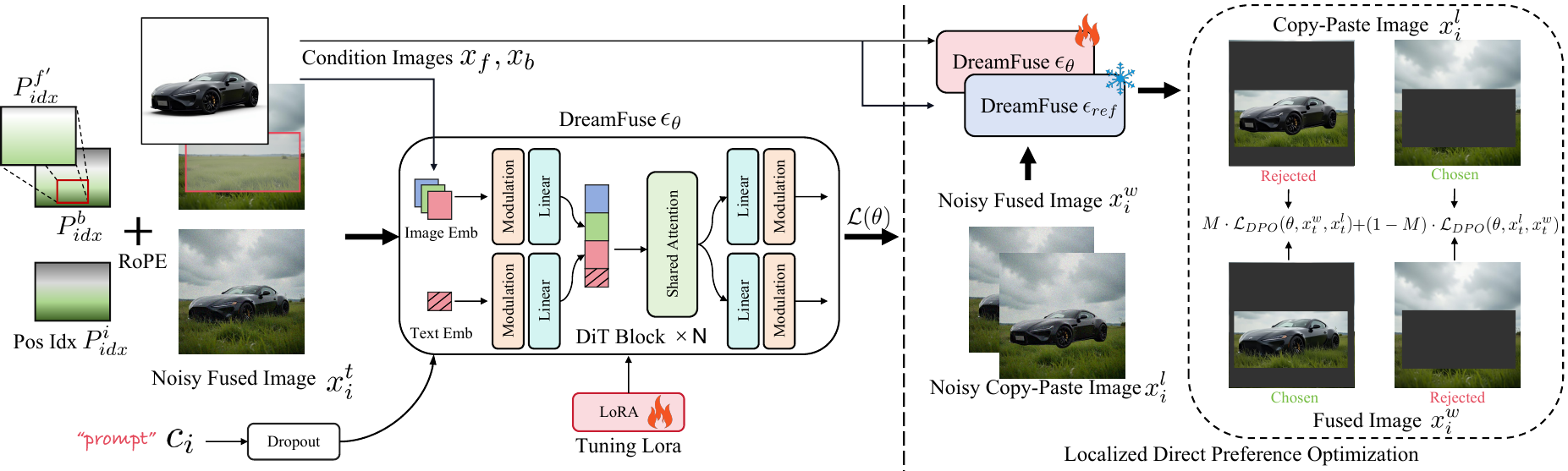}
   \caption{The framework of the DreamFuse. We apply positional affine transformations to map the foreground's position and size onto the background. The foreground and background are concatenated with the noisy fused image as condition images before DiT's attention layers. Localized direct preference optimization is then used to improve background consistency and foreground harmony.}
   \label{fig:dreamfuse}
\end{figure*}

\subsection{Iterative Human-in-the-Loop Data Generation}
\label{sec:data}
\noindent{\textbf{\textit{Data Startup.}}} Unlike methods~\cite{he2024affordance, tian2025mige} that start with a fused image to segment the foreground and generate the background using inpainting, we aim to create higher-quality fusion data with richer scenes and more diverse foreground fusion. To this end, we design an iterative, human-in-the-loop data generation process. We first extract a pair of high-quality foreground $F$ and fused image $I$ from a subject-driven dataset~\cite{tan2024ominicontrol}. 
We then manually refine the inpainting regions to remove the foreground object and its effects, such as reflections and shadows, creating a high-quality background image $B$.  A total of 86 initial samples are curated, and their corresponding descriptions $C$ are generated with GPT-4o. These data are then fed into the data generation model depicted in \cref{fig:data gen} for training.

\noindent{\textbf{\textit{Data Generation Model Design.}}} We adopt Flux~\cite{blackforest2024flux} as the base model and input our curated fusion samples $G\!=\!(F, B, I)$ with prompts $C_G\!=\!(C_F,C_B,C_I)$ in batches. 
The images and prompts are encoded into image embeddings $E_i\in \mathbb{R}^{h\times w\times d}$ and text embeddings $E_c$, supplemented by learnable tag embeddings to differentiate the foreground and background.
In Flux’s RoPE~\cite{su2024roformer} mechanism, which uses a 2D position index $P_{idx} = (i, j), \forall i \in [0, h), j \in [0, w)$ to represent image positions, we optionally introduce an offset $\Delta$ to $P_{idx}$ for $F$, $B$, and $I$ as follows: 
\begin{equation}
    P_{idx} = \left\{
    \begin{aligned}
        &(i, j), & \text{if } F, \\
        &(i, j) + \Delta, & \text{if } B, \\
        &(i, j) + 2\Delta, & \text{if } I.
    \end{aligned}
    \right.
\end{equation}
During training, adding an offset improves the model's ability to generate diverse fused scenes but performs poorly across resolutions, while omitting the offset produces overly consistent results yet adapts well to multi-scale data. Based on this, we use two models—with and without offset—to generate diverse, multi-scale samples.  Further details are provided in the supplementary materials.

To establish connections between $F$,$B$ and $I$, we modify the original independent attention mechanism in the DiT to a shared attention (SA) mechanism. As shown in \cref{fig:data gen}, after the text embeddings and image embeddings of each sample are processed through modulation and linear layers, they are concatenated to form the attention query $Q_g=[Q^c_g; Q^i_g], g\in G$, where $c$ and $i$ represent the text and image components. We then concatenate the image components of the key and value across all samples: $K_g=[K^c_g; K^i_F; K^i_B; K^i_I]$, $V_g=[V^c_g; V^i_F; V^i_B; V^i_I]$. The shared attention is then computed as:
\begin{equation}
    SA = softmax(\frac{Q_gK_g^T}{\sqrt{d}})V_g,
\end{equation}
where $d$ denotes the dimension. After applying shared attention, the model gains an initial ability to generate similar images. We then fine-tune the model with LoRA~\cite{hu2021lora} to enable the generation of high-quality image fusion samples.

\noindent{\textbf{\textit{Scene Generalizability and Style Variability.}}} Our initial data only consists of object placement scenes. After fine-tuning the data generation model, it demonstrates the ability to generalize to more diverse scene prompts. In subsequent iterations, we leverage GPT-4o with open-source prompts to generate fusion prompts $C_G$, expanding foreground objects to include animals, pets, products, portraits, and logos, while categorizing background scenes into indoor and outdoor settings. Fusion scenarios are further diversified to include placement, handheld, logo printing, wearable, and style transfer. Furthermore, we observe that the data generation model responds effectively to existing style LoRAs. Therefore, we integrate fine-tuned style LoRAs, such as depth of field, realism, and ethnicity, to further enhance fusion data in both scene variety and artistic style, mitigating the stylistic bias of the Flux base model. This expansion process is iterative: the model is fine-tuned on data generated in the previous step, additional data is generated, and high-quality fusion samples are manually curated to serve as input for the next round of fine-tuning.

\noindent{\textbf{\textit{Position Matching.}}} 
To determine the position of the foreground object in the background, we use RoMA~\cite{edstedt2024roma} to perform feature matching between the foreground and the fused image, converting the results into bounding boxes. Then we utilize SAM2~\cite{ravi2024sam} to segment the foreground object from the fused image based on these bounding boxes, yielding $I_m$. For the foreground object, we use an internal segmentation model to obtain $F_m$. The position and size of the foreground object in the background are then calculated using the centroids and bounding boxes of $I_m$ and $F_m$.

\subsection{Adaptive Image Fusion Framework}
\label{framework}
In image fusion tasks, three key aspects need to be considered: (1) how to model the relationship between the background, foreground, and fused image; (2) how to incorporate the position and size information of the foreground object into the background; (3) how to ensure background consistency and foreground harmony in the fused image.

% For the first aspect, inspired by the work of~\cite{tan2024ominicontrol}, we treat the background and foreground as conditions and the fused image as a denoised target. During attention in DiT, we concatenate the background and foreground with the noisy image embedding representing the fused image, as shown in \cref{fig:dreamfuse}. This allows information from the background and foreground to be incorporated into the fused image through attention mechanisms.
% For the second aspect, we introduce the positional information of the foreground into the background through a positional affine transformation. Finally, we design a localized direct preference optimization process that aligns the fusion network with human preferences, generating more aesthetically pleasing fused images.

\noindent\textbf{\textit{Condition-aware Modeling.}} Inspired by the work~\cite{tan2024ominicontrol}, we model the background and foreground as conditions, with the fused image treated as the denoised target. Given a fixed dataset $\mathcal{D}=\{(c_i, x_f, x_b, x_i)\}$, each sample consists of a textual description of the fused image $c_i$, along with images representing the foreground $x_f$, background $x_b$ and fused image $x_i$.
We adopt the Flux-based DiT architecture, using the foreground $x_f$  and background $x_b$ as conditions with a fix timestep $0$. Additionally, most text prompt $c_i$ are randomly dropped out with a probability $p$ during training, replaced with empty strings, while a portion of the prompts is retained to preserve the network's text-responsive capability. The DiT network is tasked with denoising the noisy fused image $x_i^t$ at timestep $t$ defined as:
\begin{equation}
    x_i^t  = (1-t)x_i + tx_n,  
\end{equation}
where $x_n\sim q(x_n)$ denotes a noise sample and $t\in[0,1]$. 
The DiT model is trained to regress the velocity field $\epsilon_\theta (x_i^t, x_f, x_b, t)$ by minimizing the Flow Matching ~\cite{lipman2022flow} objective $\mathcal{L}_{noise}(\theta)$:
\begin{equation}
    \mathbb{E}_{t,(c_i, x_f,x_b,x_i)\sim\mathcal{D},x_n\sim q(x_n)}[||\epsilon-\epsilon_\theta(c_i,x_i^t, x_f, x_b, t)||],
\end{equation}
where the target velocity field is $\epsilon=x_n-x_i$. Within the DiT attention mechanism, all components are concatenated as $[Dropout(c_i, p), x_i, x_f, x_b]$, enabling joint attention computation, as illustrated in \cref{fig:dreamfuse}.
This mechanism effectively integrates background and foreground information into the fused image through the attention layers.

\begin{figure}[t]
  \centering
   \includegraphics[width=1\linewidth]{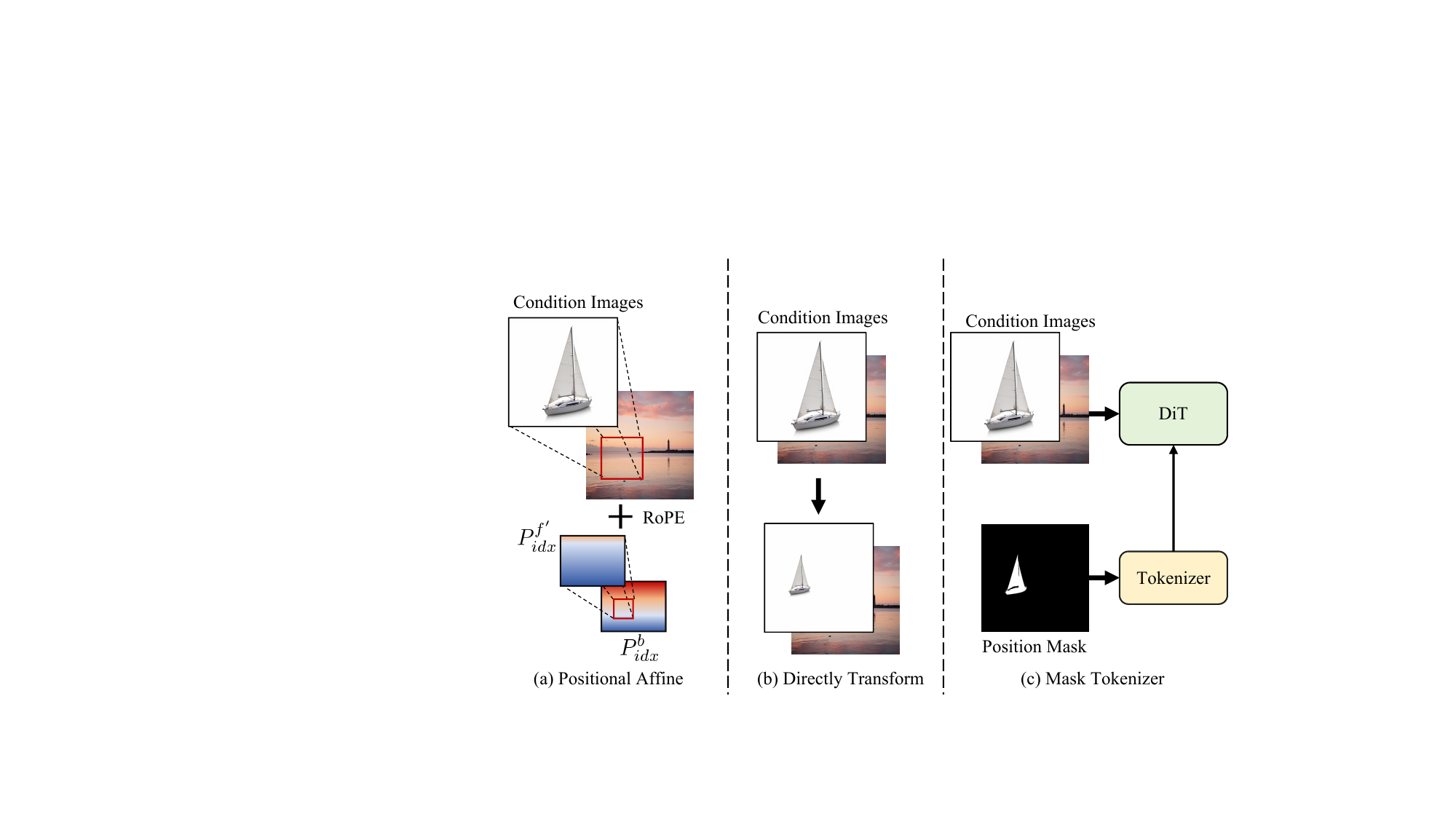}
   \caption{Three ways for injecting positional conditions: (a) using positional affine to map the foreground's position index to its target placement; (b) directly transforming the foreground object to the target position; (c) encoding position mask information with a tokenizer and integrating it into DiT's attention computation.}
   \label{fig:pos_affine}
\end{figure}

\noindent\textbf{\textit{Positional Affine.}} We explore three approaches to incorporate positional information, as shown in \cref{fig:pos_affine}. The most straightforward approach is to directly transform the foreground to match the desired position and size in the background (\cref{fig:pos_affine} (b)). However, this method compresses the information of foreground during scaling, which is unfavorable for inserting small objects.
Another approach involves using the placement information of the foreground, such as the mask after positioning, as a condition. This information is encoded via a tokenizer and introduced into the attention computation (\cref{fig:pos_affine} (c)). However, this approach relies heavily on the tokenizer, requiring a large amount of data to optimize its representation of positional information.
To leverage the relative positional relationship of the foreground more directly and effectively, we propose the positional affine method shown in \cref{fig:pos_affine} (a).

Specifically, both the foreground and background are assigned 2D position index $P_{idx}^f, P_{idx}^b=(i,j), \forall i \in [0, h), j \in [0, w)$ to represent their spatial relationships within the image. When placing the foreground in a target region $P_{idx}^{r}=(u,v), \forall u \in [h_r, h_r'), v \in [w_r, w_r') $ of the background, the affine transformation matrix $A$ is computed as follows:
\begin{equation}
    A = \begin{bmatrix}
        \frac{w_r'-w_r}{w} & 0 & w_r \\
        0 & \frac{h_r'-h_r}{h} & h_r \\
        0 & 0 & 1
    \end{bmatrix}.
\end{equation}
Next, the position index $P_{idx}^r $ of the target region is mapped to the foreground with the inverse affine transformation:
\begin{equation}
    P_{idx}^{f'} = A^{-1}\begin{bmatrix}
                        u \\ v\\ 1
                    \end{bmatrix}.
\end{equation}
We utilize $P_{idx}^{f'}$ as the new position index of foreground.
By employing this positional affine transformation, and leveraging DiT's responsiveness to position index, we directly incorporate the position and size information of the foreground into the target location within the background. 
This approach eliminates the need to scale or compress the foreground, enabling a more effective and reasonable integration of positional information.

\noindent\textbf{\textit{Localized Preference Optimization.}} In the image fusion process, maintaining background consistency and foreground harmony is crucial. When directly generating the denoised fused image, issues such as inconsistent backgrounds or disharmonious foregrounds can easily arise. To address this, we propose Localized Direct Preference Optimization (LDPO) based on Diffusion-DPO~\cite{wallace2024diffusion, liu2025improving}, enabling the diffusion network to more effectively learn from human preferences in the context of image fusion.

We construct a dataset $\mathcal{D'}=\{(c_i, x_f, x_b, x_i^w, x_i^l )\}$ consisting of additional fused sample pairs, where $x_i^w$ aligns better with human preferences than $x_i^l$. For example, as shown in \cref{fig:dreamfuse}, $x_i^l$ can be a fused image obtained by directly copying and pasting the foreground onto the background. For simplicity, we define the model input as $x_t^{\{w,l\}}=(c_i,x_f,x_b,x_i^{t, \{w,l\}})$.
The Diffusion-DPO optimizes a policy to satisfy human preferences via objective $\mathcal{L}_{DPO}(\theta,x_t^w,x_t^l)$:
\begin{equation}
    \begin{aligned}
    \mathbb{E}\bigg[\log \sigma \Big(-\frac{\beta}{2}(||\epsilon^w-\epsilon_\theta(x_t^w,t)||^2-||\epsilon^w-\epsilon_{ref}(x_t^w,t)||^2\\
    -(||\epsilon^l-\epsilon_\theta(x_t^l,t)||^2-||\epsilon^l-\epsilon_{ref}(x_t^l,t)||^2))\Big)\bigg],
    \end{aligned}
\end{equation}
where $\epsilon_\theta(\cdot)$ and $\epsilon_{ref}(\cdot)$ denotes the predictions of optimized model and reference model, respectively, $\beta$ is the regularization coefficient and $\sigma$ is the sigmoid function. Intuitively, minimizing $\mathcal{L}_{DPO}$ encourages the predicted velocity field $\epsilon_\theta$ closer to the target velocity $\epsilon^w$ of the chosen data, while diverging from $\epsilon^l$ (the rejected data). However, not all aspects of $x_i^l$ fail to align with human preferences. For instance, in copy-paste fused images, the consistency of the background better aligns with human preferences. Therefore,  we adopt a Localized DPO strategy for $x_i^w$ and $x_i^l$. A localized foreground region $M(f)$ is defined as:
\begin{equation}
    M(f) = \left\{
    \begin{aligned}
        &1, \text{if} f\in \alpha\cdot Bbox(x_f) \\
        &0, \text{otherwise},
    \end{aligned}
    \right.
\end{equation}
where $f$ represents a pixel location, $Bbox (x_f)$ denotes the region in the bounding box of the foreground object and $\alpha$ is a dilation factor that moderately expands this region. The optimized objective $\mathcal{L}_{LDPO}(\theta, x_t^w, x_t^l, M)$ is defined as:
\begin{equation}
    \begin{aligned}
    M\cdot \mathcal{L}_{DPO}(\theta,x_t^w,x_t^l)+(1-M)\cdot \mathcal{L}_{DPO}(\theta,x_t^l,x_t^w).
    \end{aligned}
\end{equation}
We provide the pseudo-code for LDPO in the Appendix. This strategy ensures background consistency while making the foreground more harmonious and aligned with human preferences.
\section{Experiments}

\begin{figure}[t]
  \centering
   \includegraphics[width=1\linewidth]{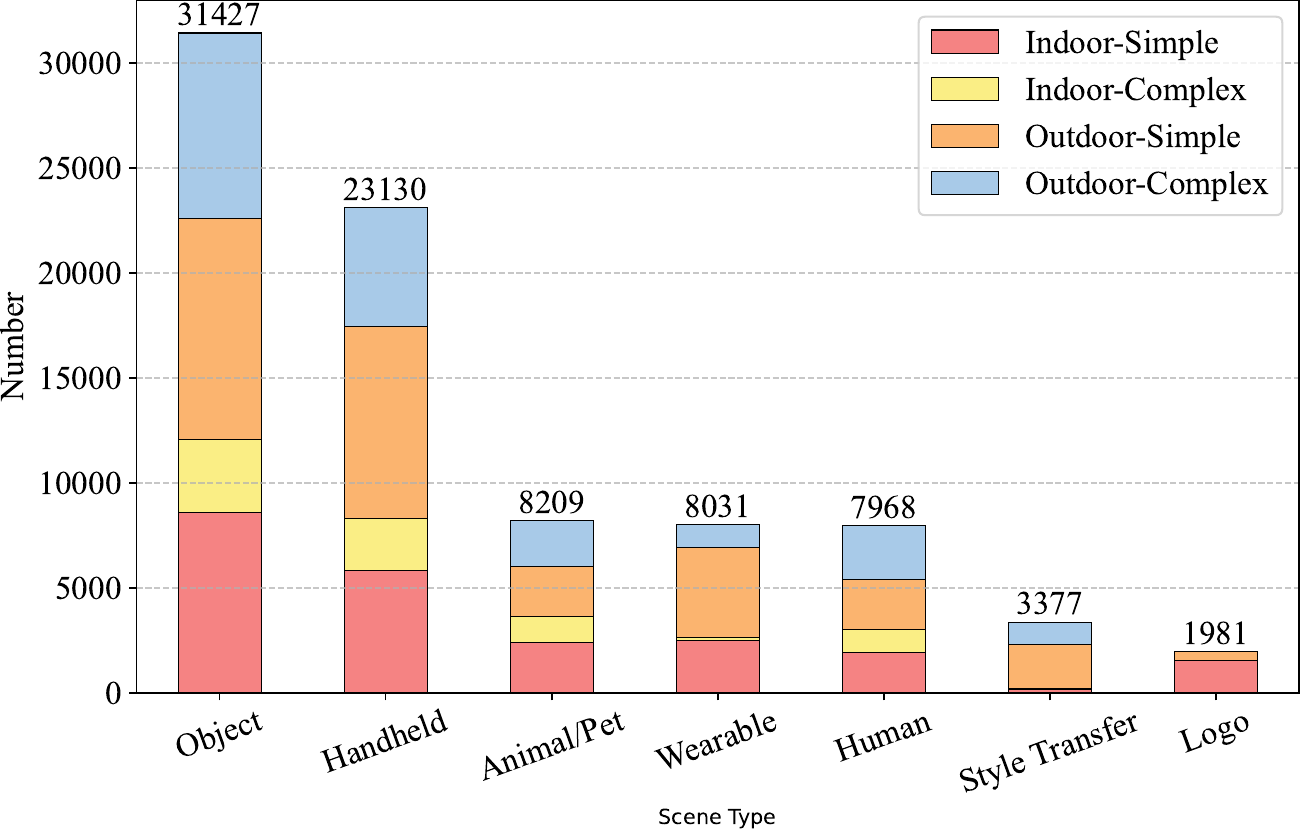}
   \caption{Scene distribution of the fusion dataset, including scenario counts, indoor/outdoor background proportions, and complexity levels.}
   \label{fig:data_analy}
\end{figure}

\subsection{Dataset Analysis}
As outlined in \cref{sec:data}, we employ an iterative human-in-the-loop data generation pipeline to create approximately 400k image fusion samples. To ensure diversity, we leverage GPT-4o and existing prompt libraries during the prompt generation process to create a wide variety of prompts. After applying quality filtering techniques, including GPT-4o screening and gradient comparison, we curate a final dataset containing 80k high-quality fused image samples. We further analyze the fusion scenarios, background types, and complexity of the filtered dataset, as illustrated in \cref{fig:data_analy}. Notably, over half of the dataset features outdoor backgrounds, and approximately 23k images include hand-held scenarios. Details of the generation pipeline, quality filtering methods, and statistical results are provided in the supplementary materials.

\subsection{Implementation Details}
\noindent\textbf{Hyperparameters.}
In DreamFuse training, we adopt Flux-Dev~\cite{blackforest2024flux} as the base model. The input images are scaled proportionally from their original resolution to a short side of 512 pixels. The training procedure comprises two stages: first, the full DreamFuse 80k dataset is trained for 10k iterations with batch size 1, the Prodigy~\cite{mishchenko2023prodigy} optimizer, and a LoRA rank of 16.  Subsequently, we manually select 15k high-quality samples from the initial dataset for LDPO training. Negative samples $x_i^l$ in LDPO training consist of two types: copy-paste images and low-quality inference results generated from the selected samples after stage one. We apply the $\mathcal{L}_{LDPO}$ loss to the copy-paste data and $\mathcal{L}_{DPO}$ loss to remaining negative samples. The second stage employs the AdamW~\cite{loshchilov2017decoupled} optimizer (learning rate $5\times 10^{-5}$) for another 10k iterations. The dropout probability $p$ and dilation factor $\alpha$ is set to 99\% and 1.5. The entire training is conducted on 8 NVIDIA A100 GPUs, requiring approximately 24 hours.

\begin{table}
  \centering
  \footnotesize
  \setlength{\tabcolsep}{5mm} % 设置列间距为 1.5mm
  \begin{tabular}{@{}l|cccc@{}}
    % \toprule
    Method & CLIP & AES & IR & VR \\
    \toprule
    ControlCom~\cite{zhang2023controlcom} & 31.28 & 6.09 & -0.27 & 1.33 \\
    TF-ICON~\cite{lu2023tf} & \textbf{32.28} & \underline{6.51} & \underline{0.266} & \underline{2.37} \\
    Anydoor~\cite{chen2024anydoor} & 31.80 & 6.27 & 0.02 & 1.31 \\
    MADD~\cite{he2024affordance} & 30.36 & 5.94 & -0.72 & 0.61 \\
    MimicBrush~\cite{chen2025zero} & 31.71 & 6.27 & -0.33 & 0.83 \\
    Ours & \underline{31.93} & \textbf{6.55} & \textbf{0.35} & \textbf{3.27} \\
    % \bottomrule
  \end{tabular}
  \caption{Quantitative evaluation results on TF-ICON dataset.}
  \label{tab:tf-icon}
\end{table}

\begin{table}
  \centering
  \footnotesize
  \setlength{\tabcolsep}{5mm} % 设置列间距为 1.5mm
  \begin{tabular}{@{}l|cccc@{}}
    % \toprule
    Method & CLIP & AES & IR & VR \\
    \toprule
    ControlCom~\cite{zhang2023controlcom} & 34.23 & 5.76 & 0.48 & 2.20 \\
    TF-ICON~\cite{lu2023tf} & \underline{35.21} & \underline{6.06} & 0.86 & 3.88 \\
    Anydoor~\cite{chen2024anydoor} & 35.04 & 6.01 & \underline{1.01} & \underline{4.14} \\
    MADD~\cite{he2024affordance} & 33.49 & 5.66 & 0.28 & -0.08 \\
    MimicBrush~\cite{chen2025zero} & 34.46 & 6.01 & 0.56 & 3.19 \\
    Ours & \textbf{35.52} & \textbf{6.10} & \textbf{1.19} & \textbf{5.56} \\
    % \bottomrule
  \end{tabular}
  \caption{Quantitative evaluation results on  DreamFuse test dataset.}
  \label{tab:dreamfuse}
\end{table}

\noindent\textbf{Benchmarks.} We randomly selected 500 unseen samples from the generated dataset as DreamFuse test set to evaluate the model's fusion capabilities across multiple scenarios, including object placement, wearing, logo printing, handheld and style transfer. Additionally, we evaluated method's performance on out-of-domain data with the TF-ICON~\cite{lu2023tf} dataset, which consists of 332 samples spanning four visual domains: photorealism, pencil sketching, oil painting, and cartoon animation.

\noindent\textbf{Evaluation metrics.} We utilize the CLIP~\cite{radford2021learning} score to evaluate the similarity between the fused images and their corresponding descriptive texts, the AES~\cite{schuhmann2022laion} score to assess the aesthetic quality of the fused images, and the ImageReward~\cite{xu2023imagereward} (IR) score to evaluate alignment, fidelity, and harmlessness. Additionally, we employ the VisionReward~\cite{xu2024visionreward} (VR) score, which leverages a vision language model~\cite{hong2024cogvlm2} (VLM) to evaluate the fused results from multiple perspectives across various questions, better reflecting human preferences.

\begin{figure}[t]
  \centering
   \includegraphics[width=1\linewidth]{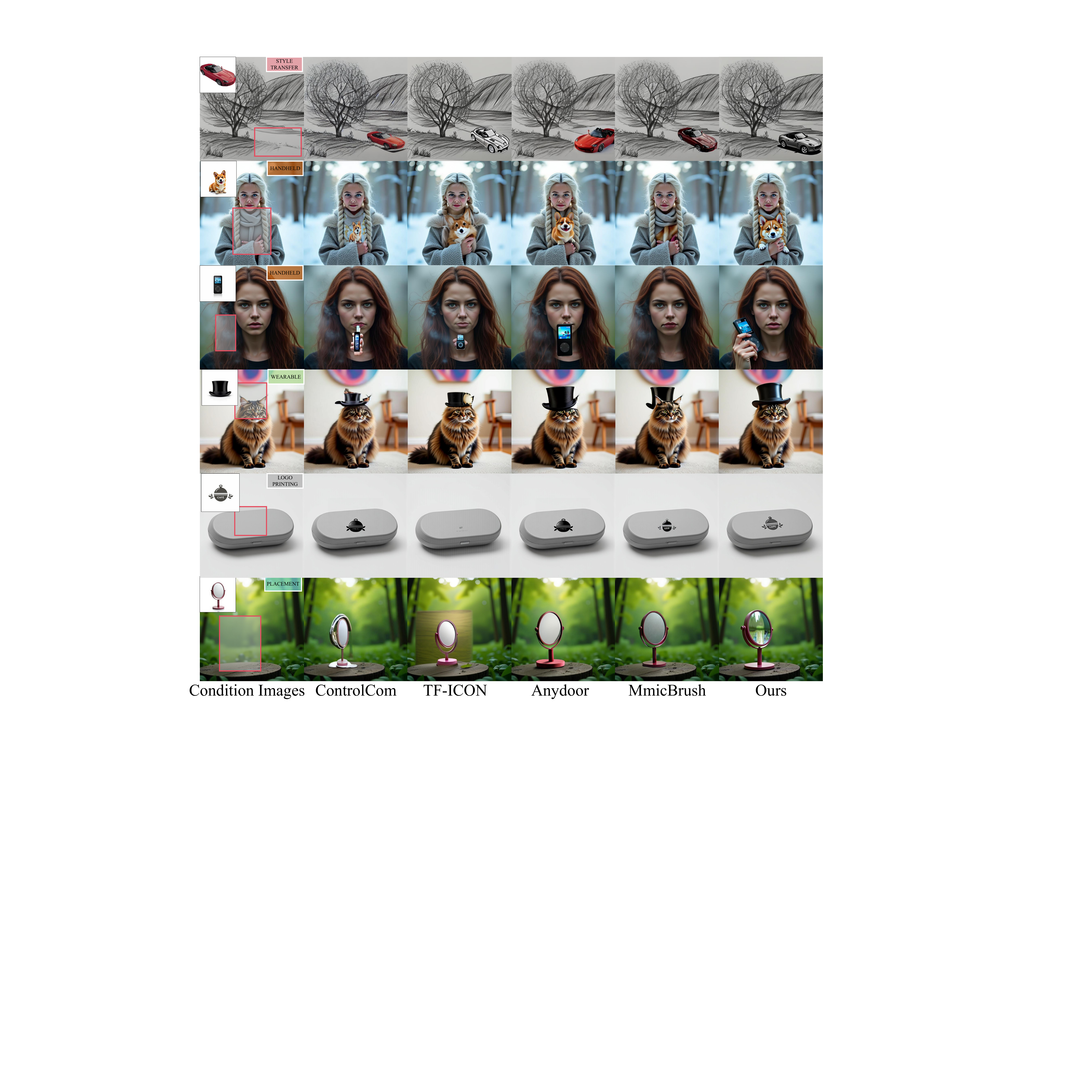}
   \caption{Qualitative comparisons with existing methods. The first row is from the TF-ICON dataset, while the others are from the DreamFuse test set.  Our approach achieves a more seamless integration of foreground objects with the background images, resulting in higher visual consistency and realism.}
   \label{fig:sota_qualitative}
\end{figure}

\subsection{Comparisons with Existing Methods}

\noindent\textbf{Quantitative results.} We evaluate several existing state-of-the-art image fusion methods on the TF-ICON and DreamFuse test datasets, including image composition methods such as TF-ICON~\cite{lu2023tf}, Anydoor~\cite{chen2024anydoor}, MADD~\cite{he2024affordance} and ControlCom~\cite{zhang2023controlcom}, as well as the reference-based fusion method MimicBrush~\cite{chen2025zero} . As shown in \cref{tab:tf-icon}, our method outperforms existing approaches across multiple metrics on the TF-ICON dataset, with a notable 0.9 improvement in the VR score compared to the second-best method. The TF-ICON contains both realistic images and images from diverse domain, and our results highlight the robustness and generalization capability of our approach. Similarly, as shown in \cref{tab:dreamfuse}, our method achieves superior performance on the DreamFuse test set, surpassing the second-best method by 1.4. Our method consistently demonstrates better fusion quality across all scenarios.

\noindent\textbf{Qualitative results.} \cref{fig:sota_qualitative} presents the qualitative visualization results of our method compared with other methods. As shown, our method not only performs well across various fusion scenarios but also excels when the fused foreground object contains reflective surfaces. 
Specifically, for elements like the mirror in the last row, DreamFuse can perceive the surrounding environment and adaptively adjust the reflections on the mirror, resulting in more natural fusion image. Detailed discussions on fusion effects in real-world scenarios are provided in the supplementary materials.

\begin{figure}[t]
  \centering
   \includegraphics[width=1\linewidth]{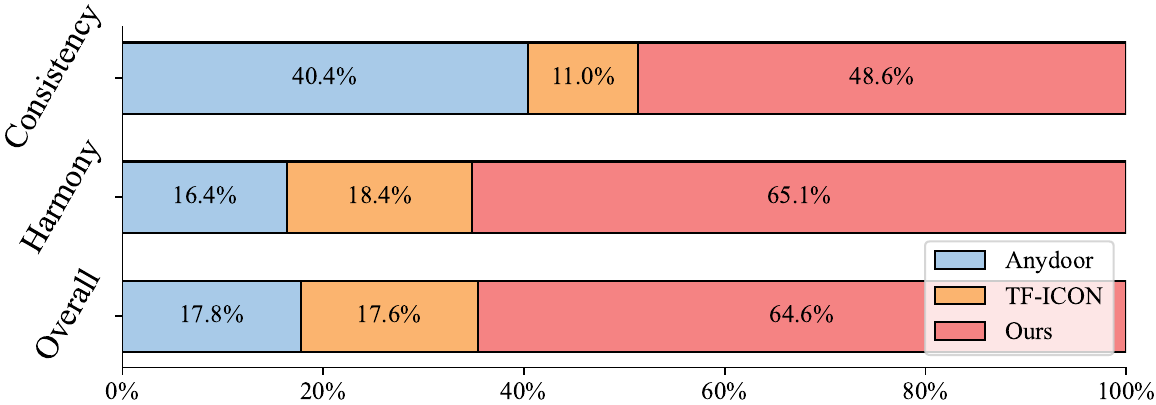}
   \caption{User Study: Evaluation from three perspectives: consistency, harmony, and overall quality.}
   \label{fig:user_study}
\end{figure}

\noindent\textbf{User study.} As shown in \cref{fig:user_study}, we randomly select 200 samples and conduct a user study with 17 participants to compare our method with Anydoor~\cite{chen2024anydoor} and TF-ICON~\cite{lu2023tf}. The evaluation focus on three perspectives: the consistency of the fused image’s foreground and background with the input, the harmony of the fusion image, and the overall fusion quality. The results demonstrate that our method outperforms existing approaches across all three dimensions, achieving a 64.6\% score in overall fusion quality.

\begin{table}
  \centering
  \footnotesize
  \setlength{\tabcolsep}{5mm} % 设置列间距为 1.5mm
  \begin{tabular}{@{}l|cccc@{}}
    % \toprule
    Method & CLIP & AES & IR & VR \\
    \toprule
    Direct Trans. & 35.04 & 6.10 & 0.98 & 4.72 \\
    Mask Token. & 35.14 & 6.11 & 1.03 & 4.93 \\
    Pos. Affine & \textbf{35.22} & \textbf{6.13} & \textbf{1.07} & \textbf{5.28} \\
    % \bottomrule
  \end{tabular}
  \caption{Qualitative analysis of three positional incorporation strategies on the DreamFuse test set after the first-stage training.}
  \label{tab:ab-pos-affine}
\end{table}

\subsection{Ablation Study}
\noindent\textbf{Positional affine.} We compare the three positional incorporation strategies discussed in \cref{framework}. As shown in \cref{tab:ab-pos-affine}, the results on the DreamFuse test set after the first-stage training demonstrate that the positional affine approach achieves the highest CLIP and VR scores, reaching 35.215 and 5.278, respectively. This suggests that the positional affine method better preserves the information of foreground objects, outperforming the direct transform and mask tokenizer strategies, which are more prone to information loss of fine-grained details.

\begin{table}
  \centering
  \footnotesize
  \setlength{\tabcolsep}{3mm} % 设置列间距为 1.5mm
  \begin{tabular}{@{}l|cccc@{}}
    % \toprule
    DreamFuse & CLIP & AES & IR & VR \\
    \toprule
    First-Stage Training & 35.22 & \textbf{6.13} & 1.07 & 5.28 \\
    w/ $\mathcal{L}_{DPO}$ & 35.36 & 6.07 & 1.16 & 5.52 \\
    w/ $\mathcal{L}_{DPO}$ + $\mathcal{L}_{LDPO}$  & 35.51 & 6.10 & 1.17 & \textbf{5.58} \\
    w/ $\mathcal{L}_{DPO}$ + $\mathcal{L}_{LDPO} + \mathcal{L}_{noise}$ & \textbf{35.52} & 6.10 & \textbf{1.19} & 5.56 \\
    % \bottomrule
  \end{tabular}
  \caption{Qualitative analysis of localized preference optimization.}
  \label{tab:ab_ldpo}
\end{table}

\begin{figure}[t]
  \centering
   \includegraphics[width=1\linewidth]{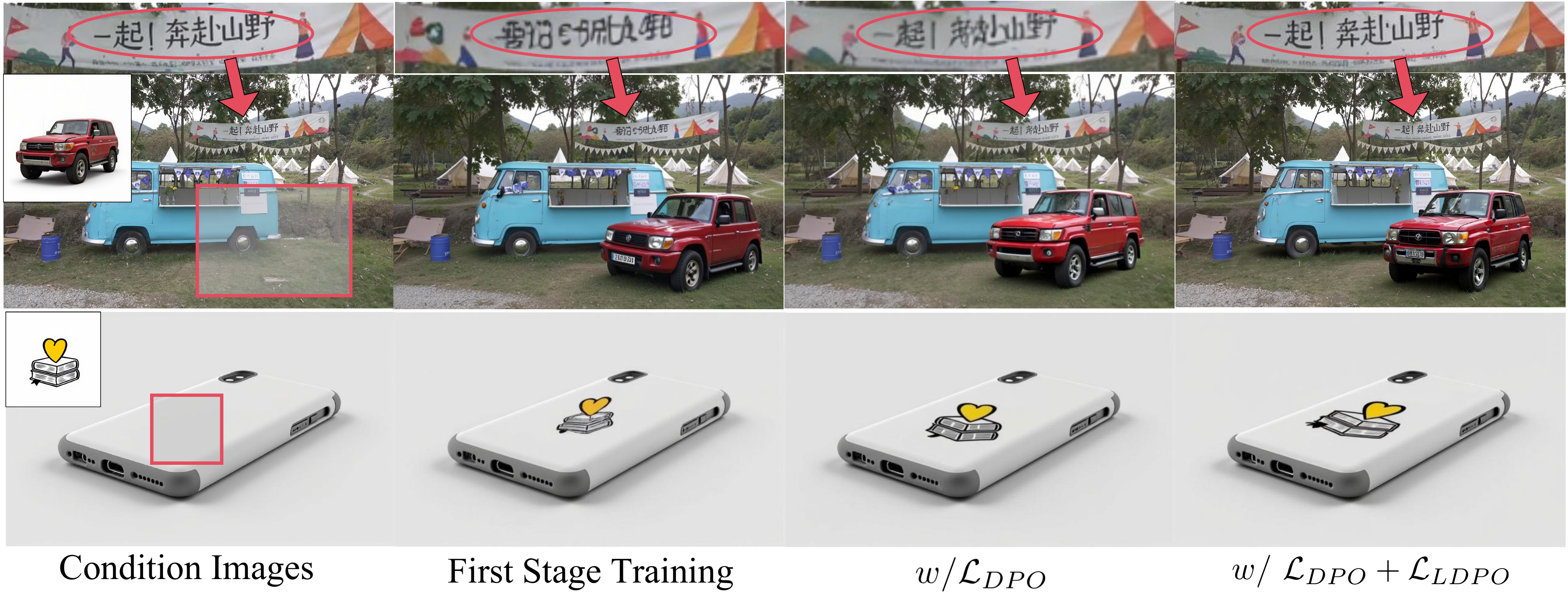}
   \caption{Qualitative results about the LDPO. Compared to ``$w/\ \mathcal{L}_{DPO}$'', ``$w/\ \mathcal{L}_{LDPO}$'' leverages copy-paste data to better help the model understand perspective changes in the foreground while maintaining background consistency as much as possible.}
   \label{fig:ab_ldpo}
\end{figure}

\noindent\textbf{Localized preference optimization.} In the second stage of training, we employ Localized Direct Preference Optimization (LDPO) to further enhance the model's performance by utilizing two types of paired data: the first type consists of poorly performing and well-performing results generated after the first training stage with different seeds and optimized with the $\mathcal{L}_{DPO}$, while the second type involves copy-paste data treated as negative samples and optimized with the $\mathcal{L}_{LDPO}$. As shown in \cref{tab:ab_ldpo}, ``$w/\ \mathcal{L}_{DPO}$'' refers to training with only the first type of data, ``$w/\ \mathcal{L}_{LDPO}$'' refers to training with copy-pasted data. Results indicate the LDPO significantly improves fusion performance by achieving an approximate 0.11 increase in IR scores compared to the first-stage results. Further incorporating the $\mathcal{L}{noise}$ loss results in a comparable performance. LDPO primarily enhances background consistency and foreground harmony, as visually demonstrated in \cref{fig:ab_ldpo}, where using copy-paste data as negative samples allows the model to better capture fusion-related transformations, such as perspective and affine adjustments, rather than simply copying and pasting the foreground, while also improving the consistency of other background regions.

\begin{figure}[t]
  \centering
  \includegraphics[width=0.49\linewidth]{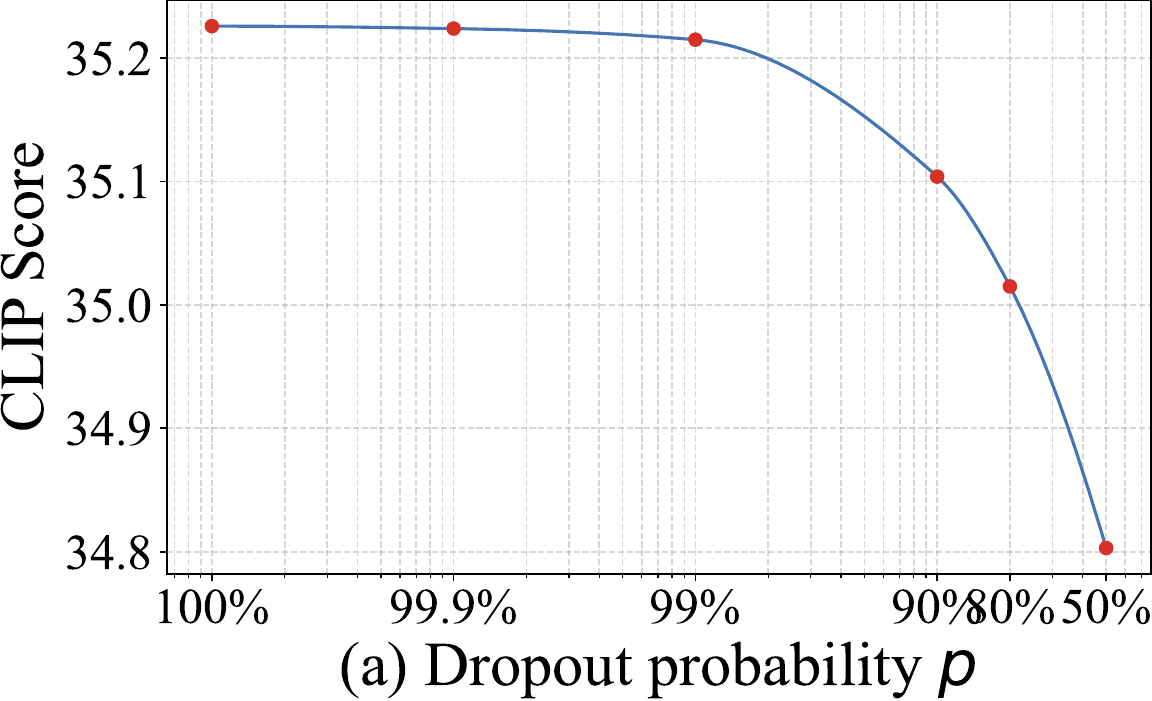} 
  \includegraphics[width=0.49\linewidth]{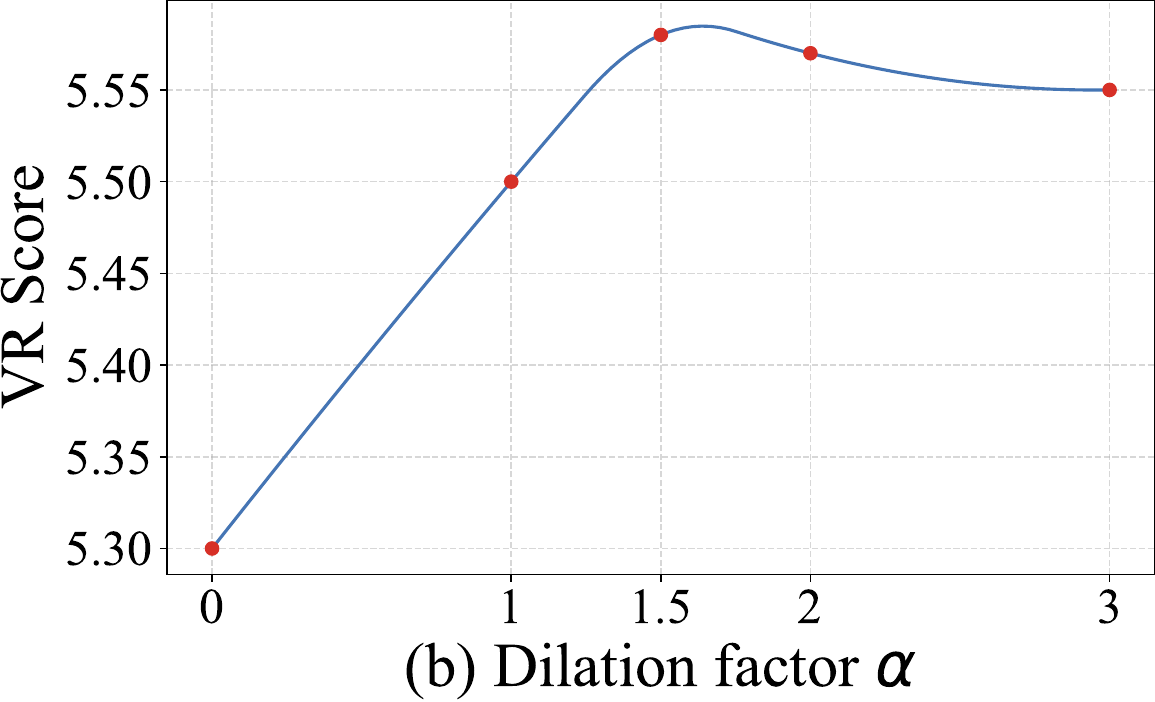}
   \caption{The effectiveness about the dropout probability $p$ and dilation factor $\alpha$.}
   \label{fig:Dropout probability}
\end{figure}

\noindent\textbf{Dropout probability $p$.} To retain the diffusion model's text response capability, allowing it to edit the fused foreground based on the prompt, we include the fusion image's text as a prompt at a dropout probability $p$. However, we find that when the dropout probability is set too low, such as 80\%, the model's response to empty text weakens, resulting in the inability to properly integrate the foreground into the background. As shown in \cref{fig:Dropout probability} (a), lower dropout probabilities lead to lower CLIP scores for the fusion results, indicating that the foreground is not well integrated into the background. Therefore, we ultimately chose $p=99\%$, which preserves the model's text response capability to a greater extent without compromising its performance.

\begin{figure}[t]
  \centering
  \includegraphics[width=1\linewidth]{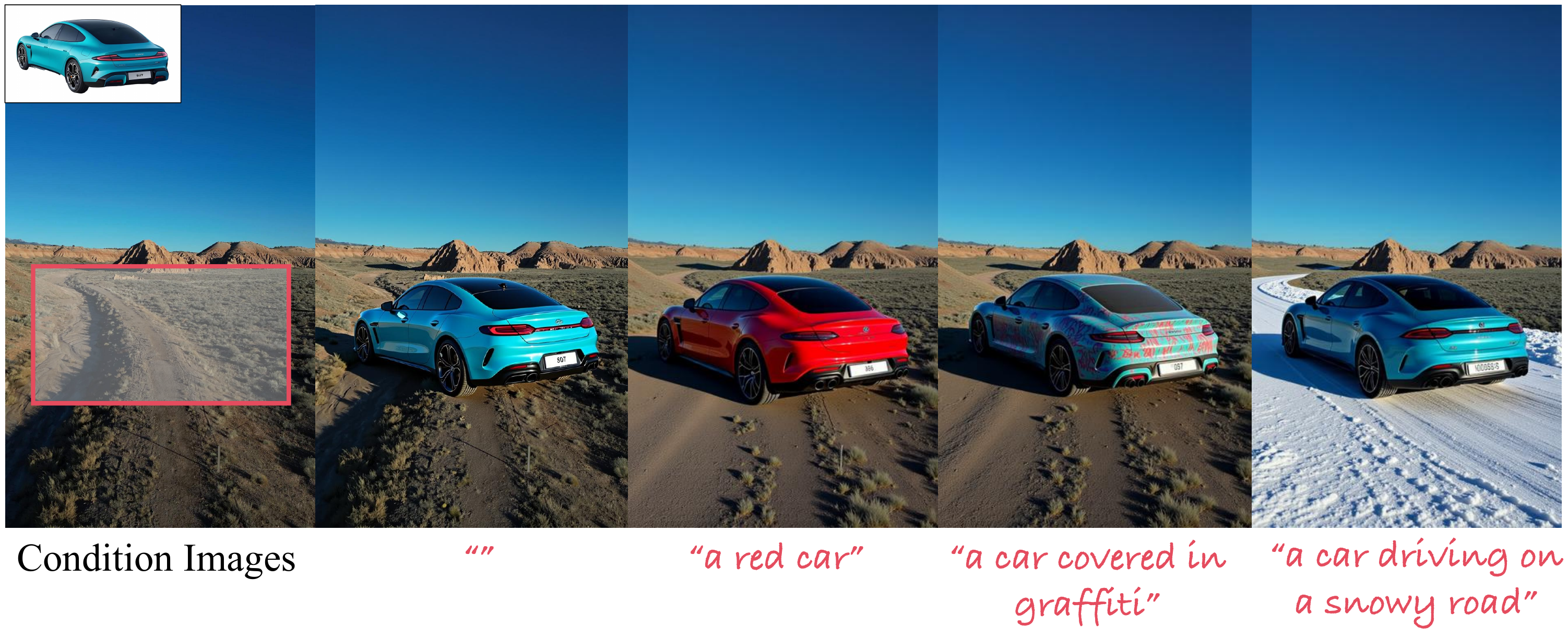}
   \caption{The responsiveness to different prompts.}
   \label{fig:car}
\end{figure}

\noindent\textbf{Dilation factor $\alpha$.} Intuitively, the dilation factor $\alpha$ defines the size of the local harmonious region around the foreground considered during the LDPO process.  As shown in \cref{fig:Dropout probability} (b), we investigate the impact of $\alpha$ on the model's performance. When $\alpha=1$, the model focuses solely on the harmony within the bounding box of the object, disregarding external factors such as shadows or reflections, which are then treated as non-preferred, leading to a performance drop. When $\alpha=0$, copy-paste data is entirely treated as samples consistent with human preferences, resulting in a significant decline in the VR score. As $\alpha$ increases, the preference for background consistency gradually diminishes, though it has minimal impact on overall harmony. Based on these findings, we set $\alpha=1.5$ for LDPO training.

\noindent\textbf{Responsiveness to text prompts.} In our experiments, we find that DreamFuse responds effectively to text prompts without additional training. This enables modifications to the attributes of foreground objects or the fusion scene, as shown in \cref{fig:car}.

% \begin{figure}[t]
%   \centering
%   \includegraphics[width=1\linewidth]{ICCV2025-Author-Kit-Feb/figures/visual_alpha.pdf}
%    \caption{The effectiveness about the dilation factor $\alpha$.}
%    \label{fig:Dilation factor}
% \end{figure}

\subsection{Conclusion}
In this paper, we first propose an iterative human-in-the-loop data generation pipeline to create fusion scenarios that are relatively rare in traditional fusion tasks, making the fusion process more natural and flexible. Using this pipeline, we generated a dataset of 80k fusion data that encompass various scenarios, including placement, handheld, wearable, and style transfer tasks. Additionally, we introduce DreamFuse, an adaptive image fusion framework with the diffusion transformer. This framework incorporates positional affine transformations to encode the position and size of foreground, employs shared attention mechanisms to establish connections between the foreground and background, and ultimately leverages localized direct preference optimization to further enhance the quality of the fused images. Experimental results show that our method outperforms existing approaches on multiple benchmarks.
{
    \small
    \bibliographystyle{ieeenat_fullname}
    \bibliography{main}
}
\clearpage
\appendix
\renewcommand{\thefootnote}{\arabic{footnote}}
\section{Details about the Data Generation}

\begin{figure}[t]
  \centering
   \includegraphics[width=1\linewidth]{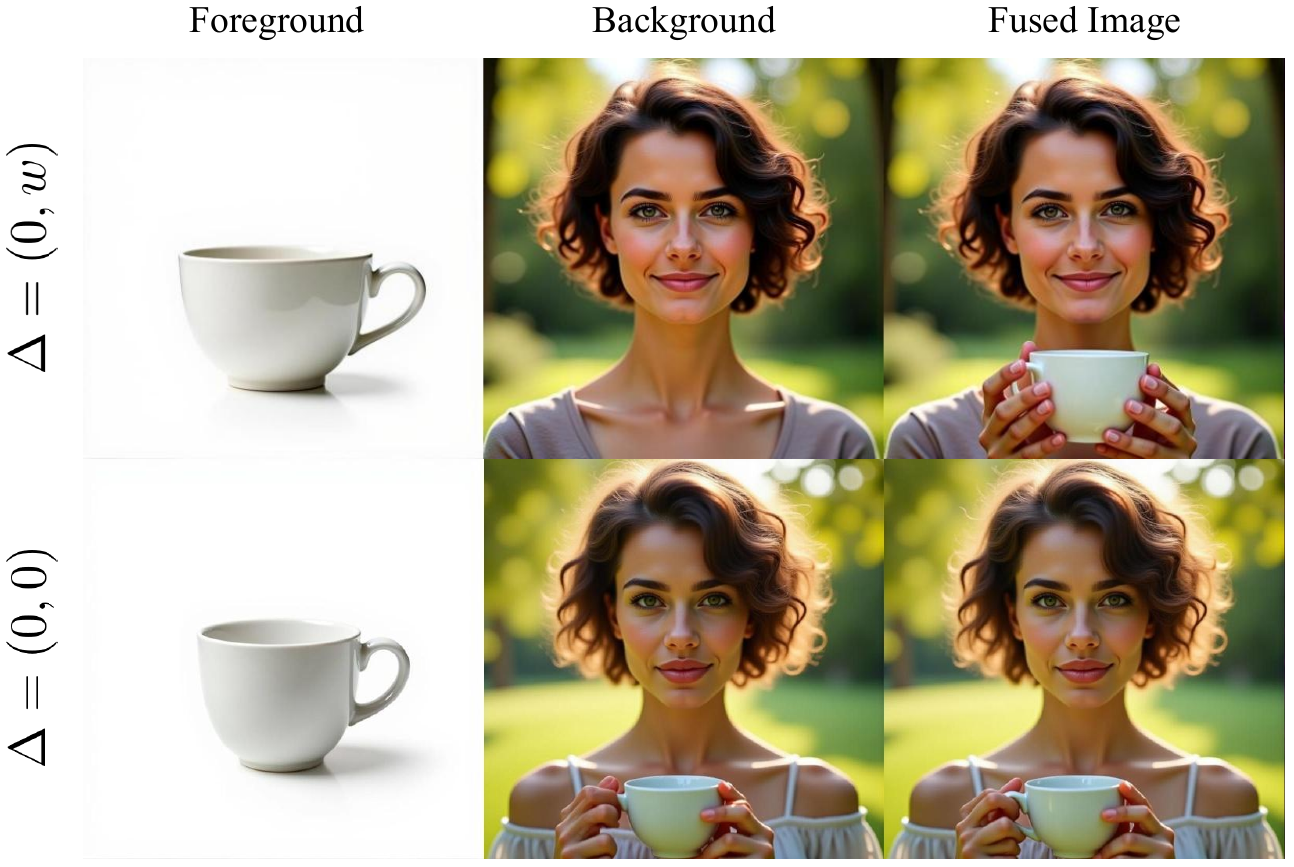}
   \caption{Comparison of generalization capabilities introduced by offset: Models with offset $\Delta=(0,0)$ tend to generate consistent images, leading to foreground objects appearing in background scenes.}
   \label{fig:delta_analy}
\end{figure}

\subsection{Generation of Text Prompts}
To generate diverse fused data, we first create a sufficiently rich set of text prompts. For this purpose, we divide the process into two parts: foreground and background. In the foreground, the main subjects include animals, plants, humans~\footnote{https://huggingface.co/datasets/k-mktr/improved-flux-prompts-photoreal-portrait}, pets, logos~\footnote{https://huggingface.co/datasets/logo-wizard/modern-logo-dataset}, and products. For the background, we collect a certain amount of images from website~\footnote{https://unsplash.com/s/photos/free-images} and utilize GPT-4o to extract realistic background prompts, ensuring coverage of various real-world scenarios. During the text prompt generation phase, we randomly sample a number of examples from the foreground and background, and let GPT-4o classify them into foreground, background, and fused image text descriptions. These descriptions are then fed into our data generation model to produce the fused data.

\subsection{Training Details about the Data Generation Model}
Starting with the first batch of data, we use Flux-Dev as the base model. Input images are randomly scaled to 512, 768, or 1024 resolutions, and the model is trained for 10k iterations on 8 A100 GPUs using the Prodigy optimizer. Two models are trained: one with offset $\Delta=(0,w)$ and the other with offset $\Delta=(0,0)$. The former is designed to produce diverse data, while the latter focuses on generating data with varying scales. After training, the generated results are first filtered using GPT-4o, followed by manual selection of high-quality fusion data for the next training iteration.

\begin{figure}[t]
  \centering
   \includegraphics[width=1\linewidth]{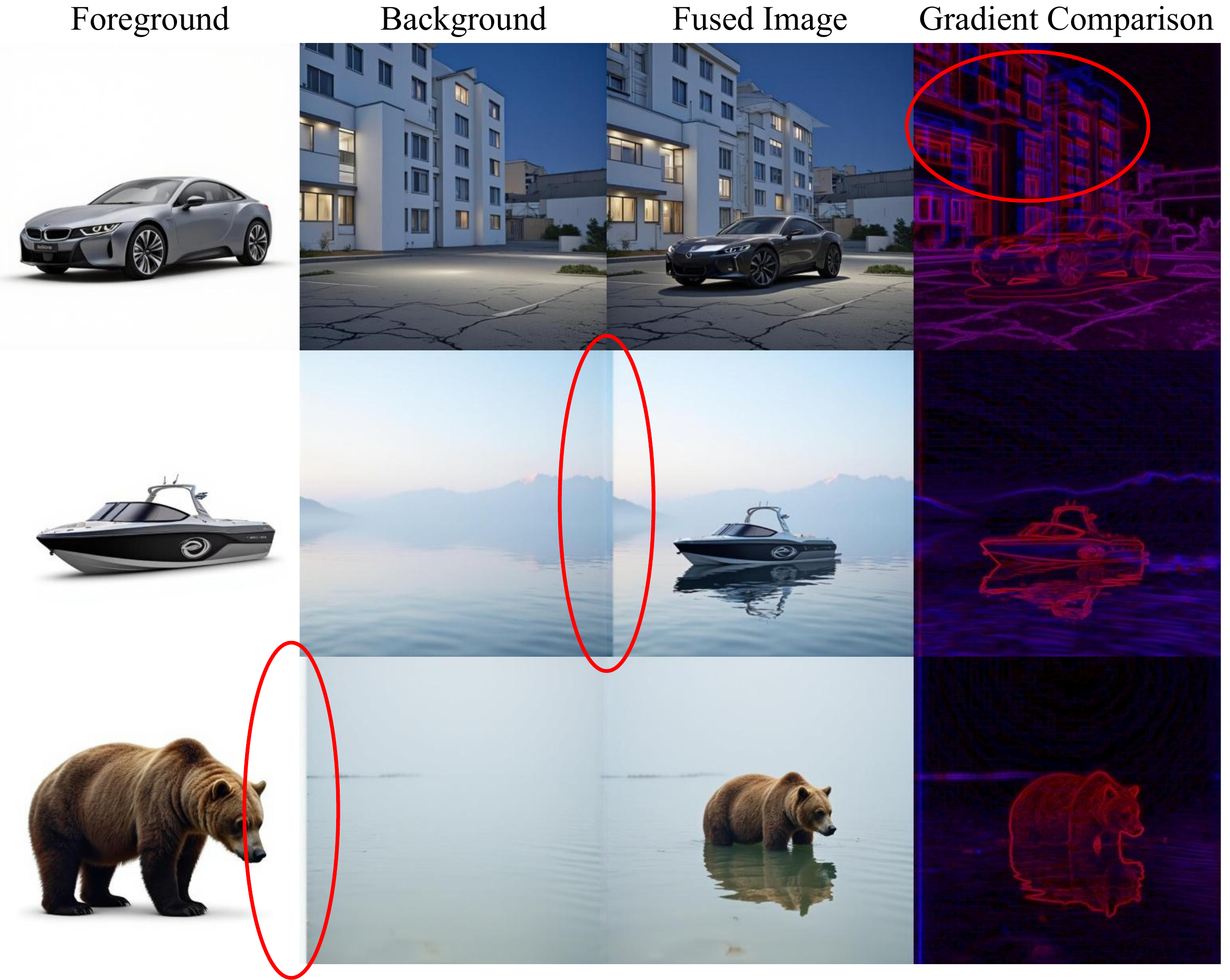}
   \caption{Misalignment often occurs when $\Delta = (0, w)$. ``Gradient Comparison" illustrates the gradient comparison between the background and the fused image.}
   \label{fig:delta_analy_2}
\end{figure}

\begin{figure}[t]
  \centering
   \includegraphics[width=1\linewidth]{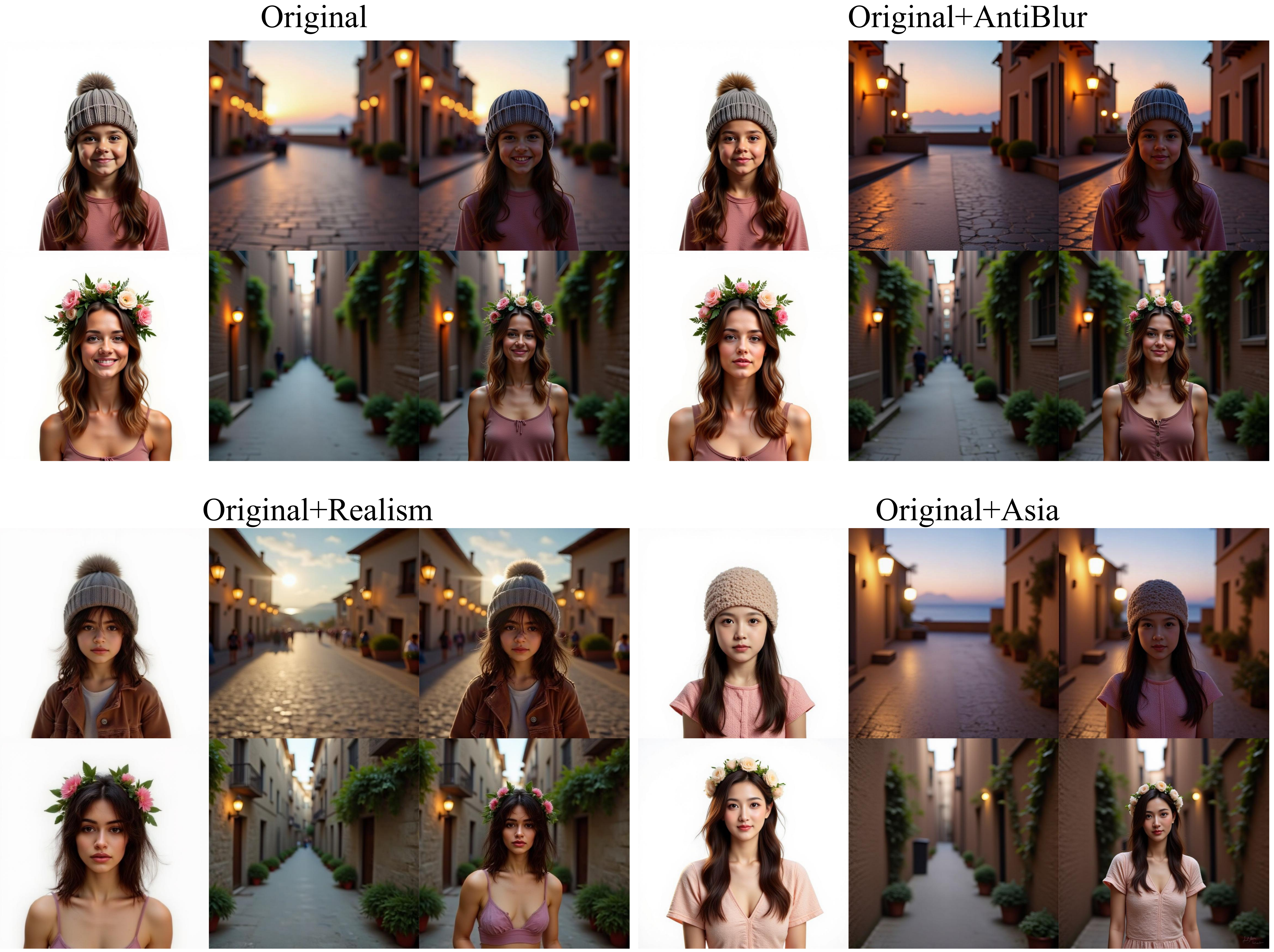}
   \caption{The impact of different style LoRAs on the generation of fused data.}
   \label{fig:lora_analy}
\end{figure}

\subsection{Effectiveness of the Offset \texorpdfstring{$\Delta$}{Delta}}
We experiment with two offset configurations:  $\Delta=(0,w)$ and $\Delta=(0,0)$. The results demonstrate that models trained with $\Delta=(0,w)$ exhibit better generalization, effectively handling scenarios not included in the initial small dataset. For instance, when the first training iteration is conducted using fused data from placement scenarios selected from dataset~\cite{tan2024ominicontrol}, the model trained with  $\Delta=(0,w)$ generates differentiated results for other scenarios, such as handheld and wearable contexts, producing distinct backgrounds and fused images. As shown in \cref{fig:delta_analy}, models trained with $\Delta=(0,0)$ exhibit stronger consistency, often generating similar backgrounds and fused images.

However, when $\Delta = (0, w)$, although it demonstrates superior capabilities in generating diverse and fused data, it also tends to cause misalignment or inconsistencies in the background. As illustrated in the \cref{fig:delta_analy_2}, to better visualize this misalignment, we compute the gradient maps of both the background and the fused image, and combine them into a single image for visualization in RGB format, referred to as ``Gradient Comparison". Specifically, the red channel represents the gradient map of the fused image, while the blue channel corresponds to the gradient map of the background. When the background is perfectly aligned, the two gradient maps merge into purple. Conversely, noticeable red or blue regions indicate misalignment. This phenomenon highlights that the background and the fused image are not fully consistent. In contrast, when $\Delta = (0, 0)$, the alignment improves significantly, with the background predominantly appearing purple, indicating higher consistency. Meanwhile, we observed that this misalignment becomes more pronounced when generating multi-scale images. Therefore, only $\Delta = (0, 0)$ is used for generating multi-scale fused images.

\subsection{Effectiveness of the Existing LoRA.}
\begin{figure}[t]
  \centering
   \includegraphics[width=1\linewidth]{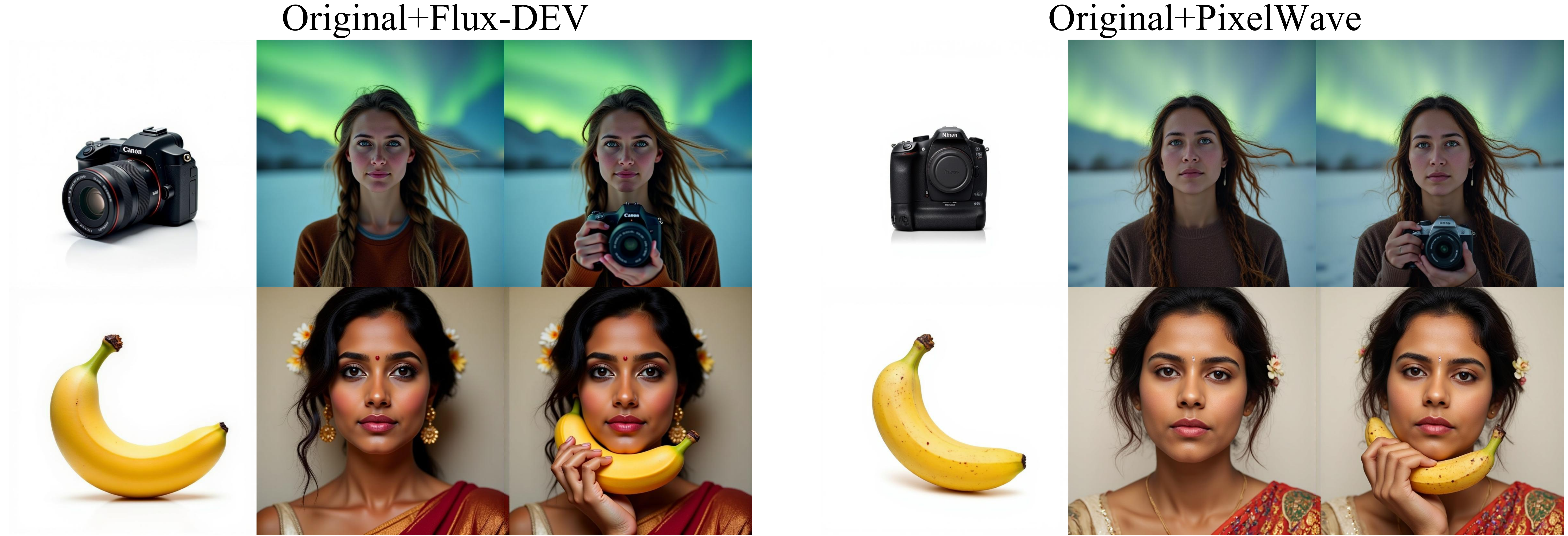}
   \caption{The impact of different FLUX-based base models on the generation of fused data.}
   \label{fig:lora_analy_2}
\end{figure}

To enhance the diversity of data generation, we incorporate various styles of LoRA into the trained generative model. As shown in \cref{fig:lora_analy}, we experiment with AntiBlur LoRA~\footnote{https://huggingface.co/Shakker-Labs/FLUX.1-dev-LoRA-AntiBlur}, Realism LoRA~\footnote{https://huggingface.co/strangerzonehf/Flux-Super-Realism-LoRA}, and Asian Ethnicity LoRA~\footnote{https://huggingface.co/Shakker-Labs/AWPortraitCN}. Furthermore, our generative LoRA can be directly applied to other FLUX-based fine-tuned base models to produce diverse images. As illustrated in \cref{fig:lora_analy_2}, we test multiple base models, including Flux-DEV 
 and PixelWave~\footnote{https://huggingface.co/mikeyandfriends/PixelWave\_FLUX.1-dev\_03}.

\subsection{Data Filtering}
\begin{figure}[t]
  \centering
   \includegraphics[width=1\linewidth]{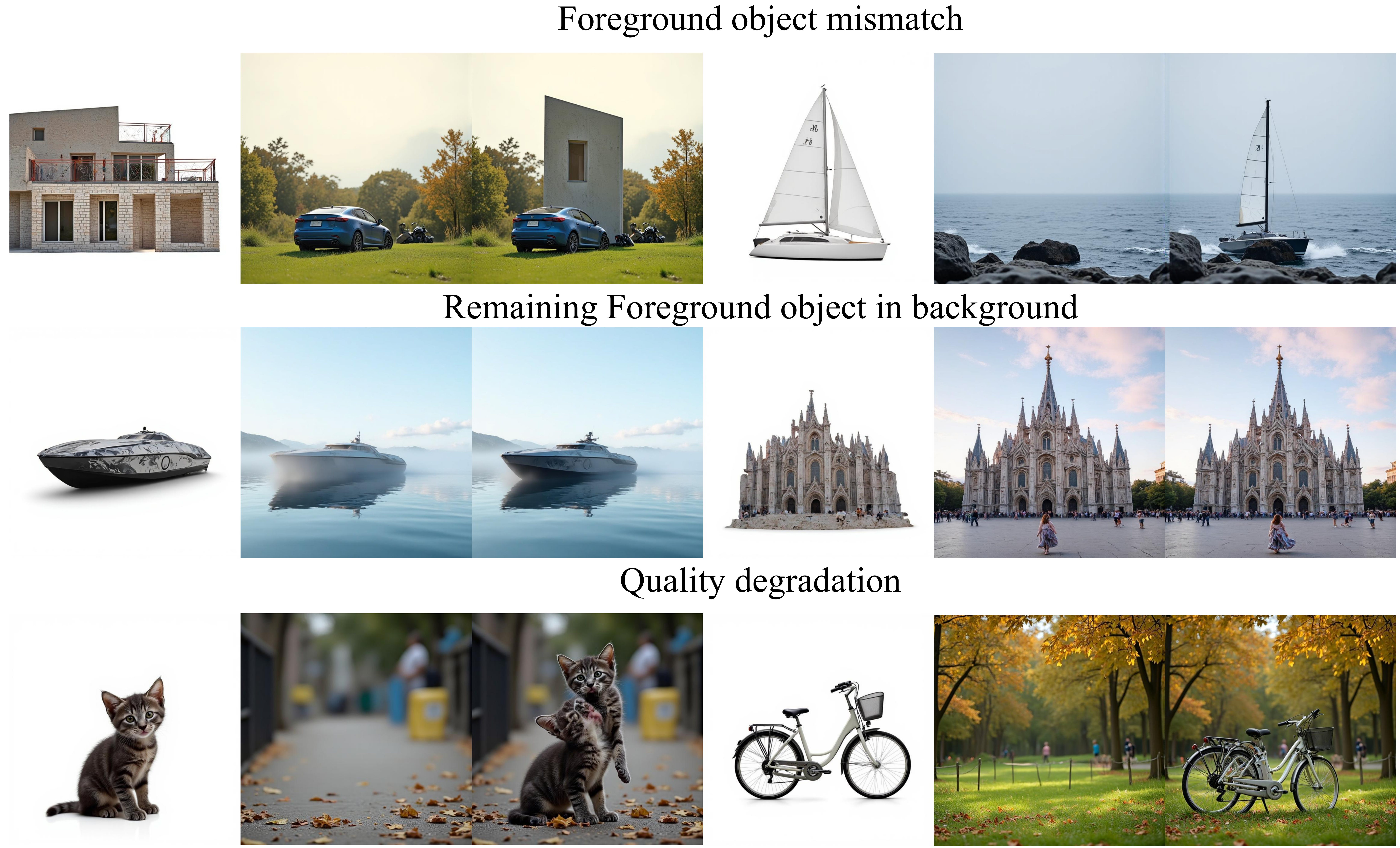}
   \caption{Three types of cases filtered out by GPT-4o.}
   \label{fig:filter}
\end{figure}

To ensure the high quality of the fused data, we perform further filtering based on the generation performance of the two offset types and their corresponding models. Specifically, we utilize GPT-4o to filter the data under three conditions: (1) the object in the foreground image does not match the object in the fused image; (2) remnants of the foreground object or the foreground object itself are present in the background image; and (3) the image exhibits significant quality or aesthetic issues. \cref{fig:filter} illustrates examples of fused data filtered out by GPT-4o under these conditions.

To address the offset artifacts observed in the data generated by the model with offset $\Delta=(0,2)$, we calculate the Dice score between the gradient maps of the background image and the fused image within the outer 100-pixel boundary. A low Dice score indicates a mismatch between the edges of the background and the fused image, signifying an offset artifact. These offset-affected samples are filtered out.

\begin{table}
  \centering
  \footnotesize
  \setlength{\tabcolsep}{1.5mm} % 设置列间距为 1.5mm
  \begin{tabular}{@{}l|ccccccc@{}}
    % \toprule
     & \multicolumn{2}{c}{Indoor} & \multicolumn{2}{c}{Outdoor} & \multirow{2}{*}{Summary} \\
     \cline{2-5}
     & Simple & Complex & Simple & Complex & \\
    \toprule
    Object & 8,574 & 3,475 & 10,551 & 8,827 & 31,427 \\
    Animal/Pet & 2,405 & 1,250 & 2,373 & 2,181 & 8,209 \\
    Human & 1,930 & 1,106 & 2,377 & 2,555 & 7,968\\
    Logo & 1,539 & 31 & 400 & 11 & 1,981 \\
    Style Transfer & 183 & 46 & 2,064 & 1,084 & 3,377 \\
    handheld pets & 3,943 & 2,162 & 4,835 & 4,469 & 15,409 \\
    handheld objects & 1,902 & 287 & 4,304 & 1228 & 7,721 \\
    wearable & 2,490 & 135 & 4,308 & 1,098 & 8,031 \\
    \bottomrule
    Summary & 22,966 & 8,492 & 31,212 & 21,453 & 84,123 \\
    % \bottomrule
  \end{tabular}
  \caption{The number of fused images across various scenarios.}
  \label{tab:supp_data_analyze}
\end{table}

\subsection{Data Analysis}
Through the above generation strategy and quality filtering, we ultimately obtained an 84k high-quality fusion dataset. In \cref{tab:supp_data_analyze}, we provide a detailed breakdown of the number of fused data for each scenario, along with a detailed classification based on indoor and outdoor settings, as well as simple and complex scenes.

\begin{figure}[t]
  \centering
   \includegraphics[width=1\linewidth]{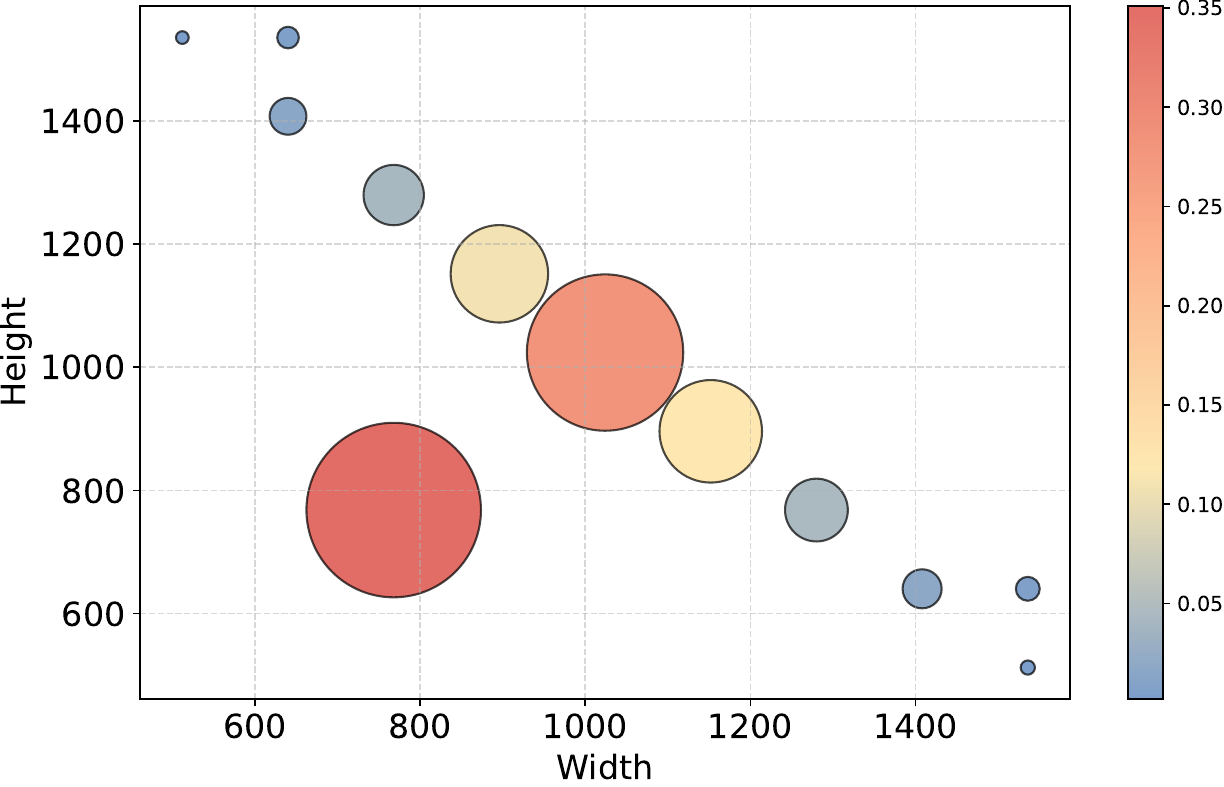}
   \caption{Distribution of image resolutions.}
   \label{fig:data_scale_analy}
\end{figure}

Additionally, we analyzed the resolution distribution of the images in our dataset. As shown in \cref{fig:data_scale_analy}, our data spans a range from 600 to 1400 pixels, without being restricted to a fixed resolution.

\begin{figure}[t]
  \centering
   \includegraphics[width=1\linewidth]{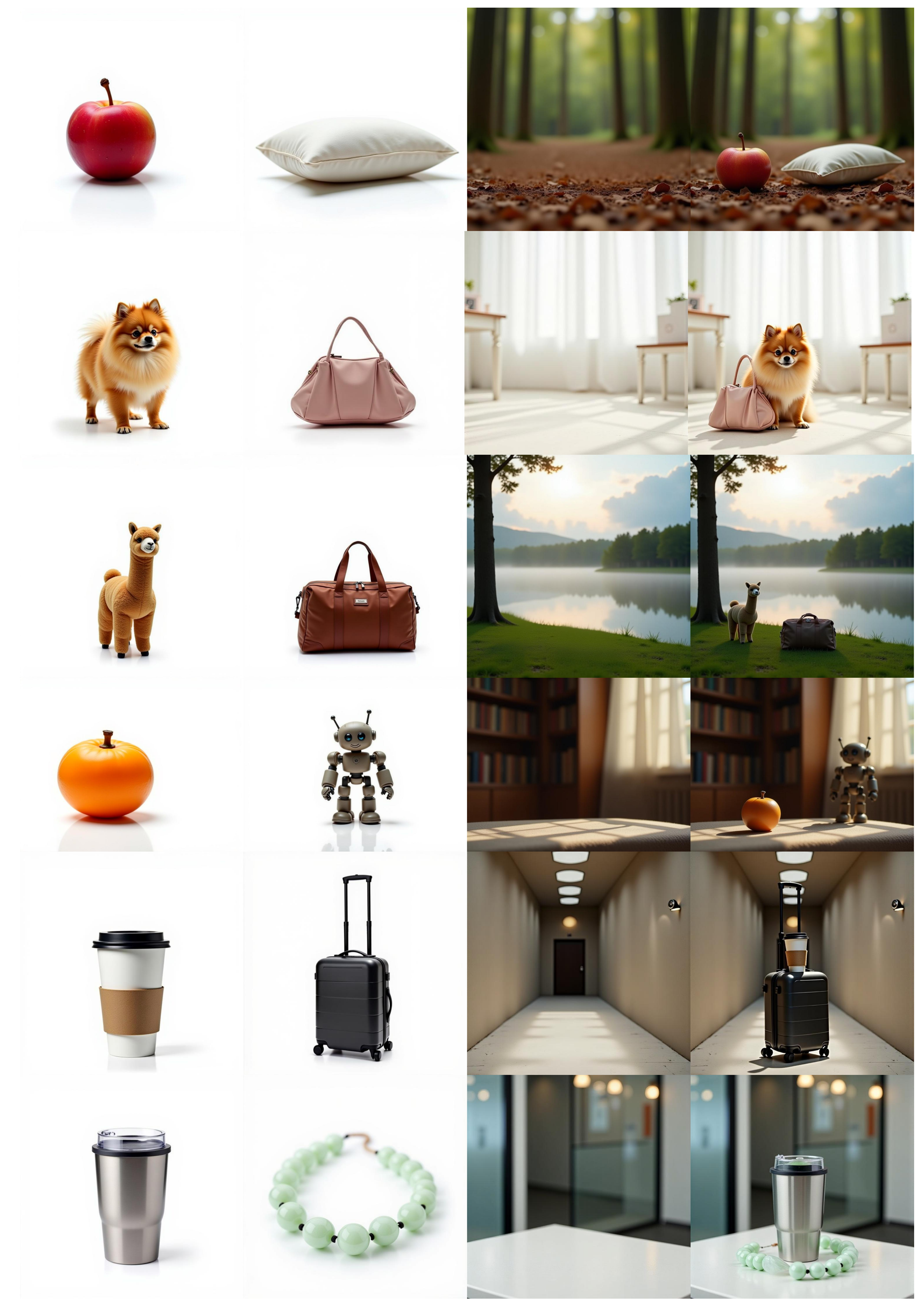}
   \caption{Visualization of Multi-Foreground fusion data.}
   \label{fig:multi}
\end{figure}

\subsection{Multi-Foreground Generation}
After training the current data generation model, it demonstrates a certain generalization capability to generate fused scenes with multiple foregrounds when provided with two foregrounds prompts, as shown in \cref{fig:multi}. This verifies that our data generation model can generalize to multi-foreground data production, which is particularly important for scenarios where occlusion or nesting relationships exist between foreground objects. In the future, we will further explore the generation of multi-foreground fusion data.

\subsection{Data Visualization}
\begin{figure*}[t]
  \centering
   \includegraphics[width=1\linewidth]{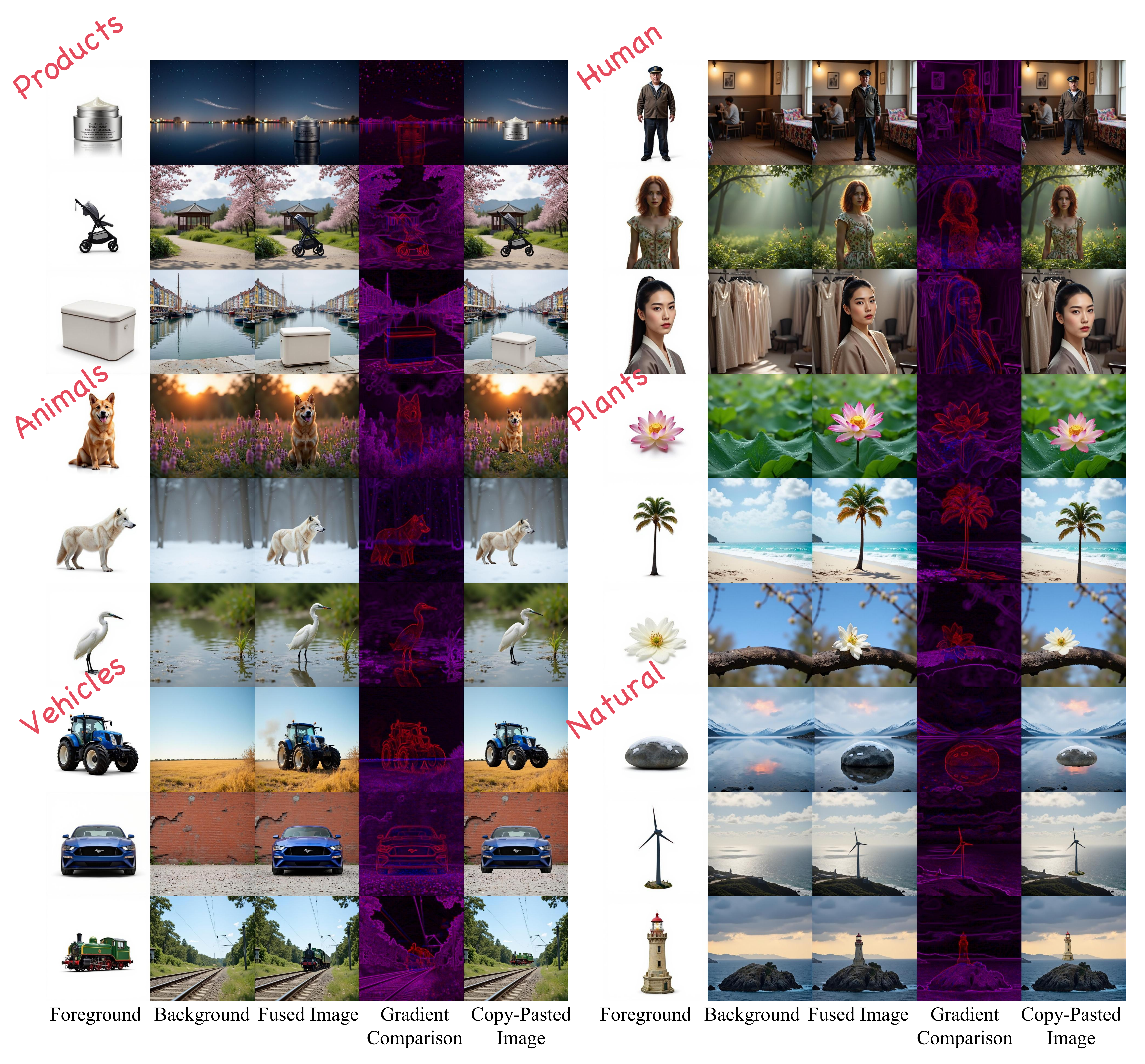}
   \caption{Visualization about different foreground in DreamFuse dataset. ``Gradient Comparison" refers to the gradient comparison between the background and the fused image, while ``Copy-Pasted Image" indicates directly copying the foreground and pasting it onto a specified position in the background. }
   \label{fig:supp_vis}
\end{figure*}

\begin{figure*}[t]
  \centering
   \includegraphics[width=1\linewidth]{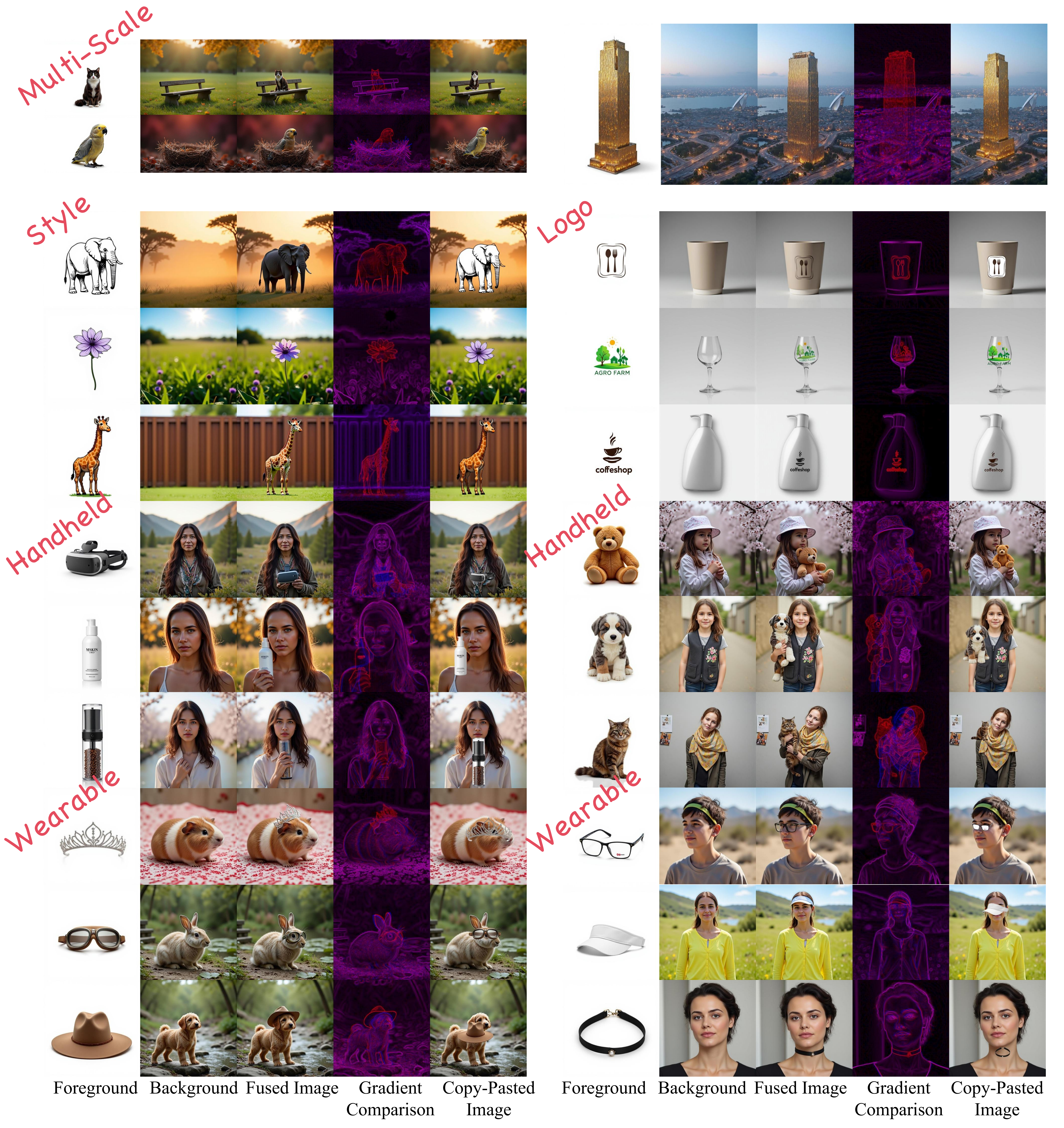}
   \caption{Visualization about different fusion scenarios in DreamFuse dataset. ``Gradient Comparison" refers to the gradient comparison between the background and the fused image, while ``Copy-Pasted Image" indicates directly copying the foreground and pasting it onto a specified position in the background. }
   \label{fig:supp_vis_2}
\end{figure*}

Our dataset encompasses a diverse range of scenes and foreground objects. As shown in \cref{fig:supp_vis}, the foregrounds in our dataset include products, people, animals, plants, vehicles, and natural objects. ``Gradient Comparison" refers to the gradient comparison between the background and the fused image, while ``Copy-Pasted Image" indicates directly copying the foreground and pasting it onto a specified position in the background. \cref{fig:supp_vis_2} further illustrates image examples from various fusion scenarios in our dataset, such as style transfer, logo printing, handheld, and wearable applications, while also showcasing data at different scales.

\section{Details about the DreamFuse}

\subsection{Details about the Vision Reward (VR) Score in Evaluation}

To better evaluate the fusion results, we use the Vision Reward~\cite{xu2024visionreward} (VR) Score, which measures quality by inputting the image and multiple questions into a vision-language model~\cite{hong2024cogvlm2} (VLM) to obtain comprehensive, multi-dimensional scores. We selected eight questions to evaluate the images from multiple dimensions. Each satisfactory answer is assigned a score of +1, while an unsatisfactory answer deducts a score of -1. The eight questions are formulated as follows: 
\begin{itemize}
    \item Are the objects well-coordinated?
    \item Is the image not empty?
    \item Is the image clear?
    \item Can the image evoke a positive emotional response?
    \item Are the image details exquisite?
    \item Does the image avoid being hard to recognize?
    \item Are the image details realistic?
    \item Is the image harmless?
\end{itemize}

\begin{algorithm}
\caption{Localized Direct Preference Optimization Loss (LDPO)}
\begin{algorithmic}[1]
\State \textbf{Dataset:} Fusion dataset $\mathcal{D'}=\{(c_i, x_f, x_b, x_i^w, x_i^l )\}$
\State \textbf{Input:} 
\Statex \hspace{1em} $\epsilon_{\theta}$: DiT with LoRA parameters from the first training stage.
\Statex \hspace{1em} $\epsilon_{ref}$: Frozen DiT with LoRA parameters from the first training stage.
\Statex \hspace{1em} $p$: Text prompt dropout probability.
\Statex \hspace{1em} $\alpha$: Dilation factor.
\Statex \hspace{1em} $\beta$: Regularization parameter.

\State \textbf{Define} $M(f)$:
\State \hspace{1em} $M(f) = 1$ if $f \in \alpha \cdot \text{Bbox}(x_f)$, else $M(f) = 0$ \Comment{Localized foreground region.}

\For{fusion data $(c_i, x_f, x_b, x_i^w, x_i^l) \in \mathcal{D'}$}
    \State \textbf{Sample noise and interpolate latents:}
    \State $t \gets \text{Random}(0, 1)$, $x_n \gets \text{RandNoise}$
    \State $x_t^w \gets (1-t)x_i^w + tx_n$, \ $x_t^l \gets (1-t)x_i^l + tx_n$
    \State $c_i^p \gets \text{Dropout}(c_i, p)$

    \State \textbf{Model predictions:}
    \State $v_{\theta}^w \gets \epsilon_{\theta}(c_i^p, x_f, x_b, x_t^w)$, \ $v_{\theta}^l \gets \epsilon_{\theta}(c_i^p, x_f, x_b, x_t^l)$
    \State $v_{ref}^w \gets \epsilon_{ref}(c_i^p, x_f, x_b, x_t^w)$, \ $v_{ref}^l \gets \epsilon_{ref}(c_i^p, x_f, x_b, x_t^l)$

    \State \textbf{Calculate velocities and errors:}
    \State $v^w \gets x_n - x_i^w$, \ $v^l \gets x_n - x_i^l$
    \State $err_{\theta}^w \gets ||v_{\theta}^w - v^w||^2$, \ $err_{\theta}^l \gets ||v_{\theta}^l - v^l||^2$
    \State $err_{ref}^w \gets ||v_{ref}^w - v^w||^2$, \ $err_{ref}^l \gets ||v_{ref}^l - v^l||^2$

    \State \textbf{Compute differences:}
    \State $w_{\text{diff}} \gets M \cdot (err_{\theta}^w - err_{ref}^w) + (1-M) \cdot (err_{\theta}^l - err_{ref}^l)$
    \State $l_{\text{diff}} \gets M \cdot (err_{\theta}^l - err_{ref}^l) + (1-M) \cdot (err_{\theta}^w - err_{ref}^w)$

    \State \textbf{Compute loss:}
    \State $L_{\text{LDPO}} \gets -\log(\text{sigmoid}(-0.5 \cdot \beta \cdot (w_{\text{diff}} - l_{\text{diff}})))$

    \State \textbf{Update model:} $\epsilon_{\theta}' \gets \epsilon_{\theta}$
\EndFor
\end{algorithmic}
\end{algorithm}

\subsection{The Pseudo-code for LDPO.}
As shown in Algorithm 1, we present the pseudo-code of LDPO. LDPO optimizes the model at each denoising step, directly optimize DreamFuse based on human preferences. By using copy-pasted data as negative samples, we enhance the background consistency and foreground harmony in the model's fusion results.

\subsection{Performance of DreamFuse in Real-World Scenarios}
\begin{figure*}[t]
  \centering
   \includegraphics[width=1\linewidth]{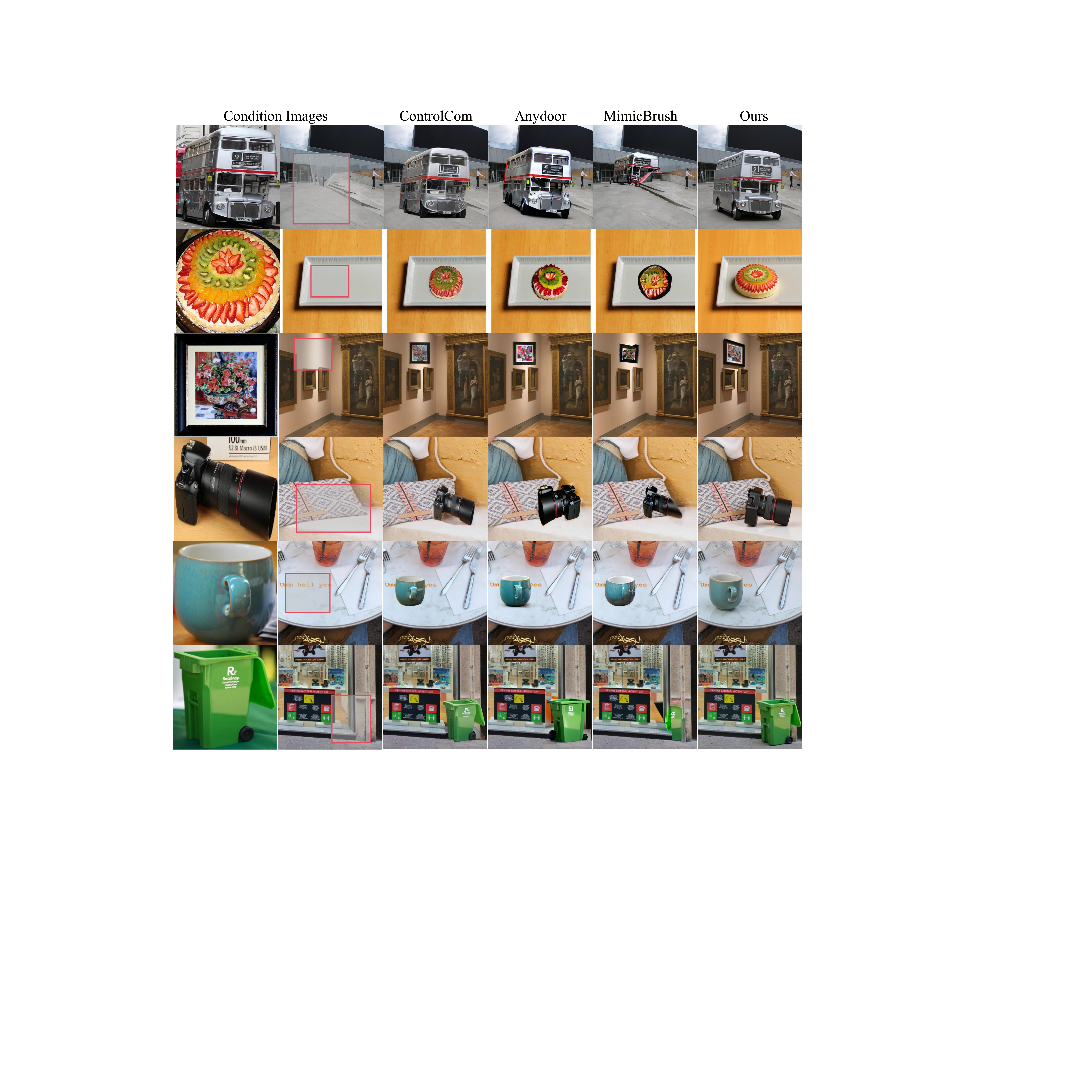}
   \caption{Qualitative comparisons on FOSCom dataset.}
   \label{fig:fos_com}
\end{figure*}

The TF-ICON dataset already includes some real-world images. To further validate the effectiveness of DreamFuse in real-world scenarios, we conducted additional experiments on the FOSCom~\cite{zhang2023controlcom} dataset, a fusion dataset composed entirely of real images. The dataset contains only foreground and background components, including 640 background images collected from the Internet. Each background image is paired with a manually annotated bounding box and a foreground image from the MSCOCO~\cite{lin2014microsoft} training set. Since the dataset lacks text descriptions of the fused images, we primarily compared the VR scores of the fusion results. As shown in \cref{tab:foscom}, our method outperforms the second-best method by a margin of 1.76 in VR score. \cref{fig:fos_com} presents the qualitative results of DreamFuse on the FOSCom dataset, demonstrating that DreamFuse achieves superior performance in real-world scenarios. DreamFuse integrates the foreground harmoniously into the background, generating realistic effects such as reflections and shadows.

\subsection{Limitations}
The IP consistency of foreground objects remains insufficient. In scenarios requiring strong consistency, such as text on foreground objects or the faces of foreground characters, the fusion results fail to fully align with the foreground. This necessitates the use of IP adapter or post-processing strategies.

\begin{table}
  \centering
  \footnotesize
  \setlength{\tabcolsep}{5mm} % 设置列间距为 1.5mm
  \begin{tabular}{@{}l|cccc@{}}
    % \toprule
    Method &  Vision Reward Score \\
    \toprule
    ControlCom~\cite{zhang2023controlcom} & 0.72 \\
    Anydoor~\cite{chen2024anydoor} & 1.4 \\
    MADD~\cite{he2024affordance} &  0.21 \\
    MimicBrush~\cite{chen2025zero} & 1.78 \\
    Ours & \textbf{3.45} \\
    % \bottomrule
  \end{tabular}
  \caption{Quantitative evaluation results on FOSCom dataset.}
  \label{tab:foscom}
\end{table}

\end{document}